\documentclass[10pt,twocolumn,letterpaper]{article}

\usepackage[pagenumbers]{iccv} 

\definecolor{iccvblue}{rgb}{0.21,0.49,0.74}
\usepackage[pagebackref,breaklinks,colorlinks,allcolors=iccvblue]{hyperref}

\usepackage{multirow}
\usepackage{color} 
\usepackage{colortbl}

\usepackage{graphicx}
\usepackage{subcaption}
\usepackage{pifont}
\usepackage{bbding}
\usepackage{graphicx}

\usepackage{array}

\def\paperID{3716} 
\def\confName{ICCV}
\def\confYear{2025}

\title{OcRFDet: Object-Centric Radiance Fields for Multi-View 3D Object Detection in Autonomous Driving}

\author{Mingqian Ji$^{1,2}$ \quad 
Jian Yang$^{1,2}$ \quad 
Shanshan Zhang$^{1,2}$ \thanks{Corresponding author.} \\
$^1$PCA Lab, Key Lab of Intelligent Perception and Systems for High-Dimensional Information of \\ Ministry of Education \\
$^2$Jiangsu Key Lab of Image and Video Understanding for Social Security, School of Computer \\ Science and Engineering, Nanjing University of Science and Technology
}

\begin{document}

\maketitle
\begin{abstract}
Current multi-view 3D object detection methods typically transfer 2D features into 3D space using depth estimation or 3D position encoder, but in a fully data-driven and implicit manner, which limits the detection performance. Inspired by the success of radiance fields on 3D reconstruction, we assume they can be used to enhance the detector's ability of 3D geometry estimation. However, we observe a decline in detection performance, when we directly use them for 3D rendering as an auxiliary task. From our analysis, we find the performance drop is caused by the strong responses on the background when rendering the whole scene. To address this problem, we propose object-centric radiance fields, focusing on modeling foreground objects while discarding background noises. Specifically, we employ Object-centric Radiance Fields (OcRF) to enhance 3D voxel features via an auxiliary task of rendering foreground objects. We further use opacity - the side-product of rendering- to enhance the 2D foreground BEV features via Height-aware Opacity-based Attention (HOA), where attention maps at different height levels are generated separately via multiple networks in parallel. Extensive experiments on the nuScenes validation and test datasets demonstrate that our OcRFDet achieves superior performance, outperforming previous state-of-the-art methods with 57.2$\%$ mAP and 64.8$\%$ NDS on the nuScenes test benchmark. Code will be available at https://github.com/Mingqj/OcRFDet.
\end{abstract}    
\section{Introduction}
\label{sec:intro}

    \begin{figure}[t!]
      \centering
        \includegraphics[width=8.2cm]{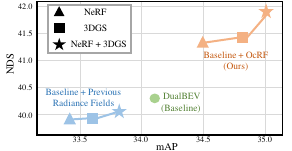}
        \caption{Comparison of previous radiance field methods directly applied to the baseline and our proposed OcRF applied to the baseline. Previous methods, whether using a single or joint field, consistently degrade performance, while our method enhances it, achieving greater improvements in the joint field setting.}
        \label{previous performance}
    \end{figure}
Multi-view 3D object detection plays a crucial role in autonomous driving. This task involves predicting object categories and their 3D coordinates based solely on multi-view RGB images. Currently, most detectors in this field project image features to the Bird's Eye View (BEV) space, and thus the main challenge lies in estimating 3D geometry within the BEV space using only RGB images. To address this problem, previous methods have made great efforts to learn effective geometric transformations from 2D to 3D space, but in a fully data-driven and implicit manner, which may limit performance improvement to some extent. Therefore, in this work, we aim to explore auxiliary tasks that are helpful to enhance the network's 3D geometry estimation capabilities, so as to facilitate multi-view 3D object detection.


Recently, radiance field methods, such as Neural Radiance Field (NeRF) \cite{nerf} and 3D Gaussian Splatting (3DGS) \cite{3DGS}, have shown strong geometry estimation capabilities in 3D reconstruction \cite{blocknerf,nerfwild,mapnerf,emernerf} and simulation \cite{drivinggaussian,streetgaussian,hugs,nerflidar} using knowledge-based geometric constraints and physical priors. Inspired by their success in occupancy prediction \cite{uniocc,renderocc,hybridocc,gaussianformer} and BEV segmentation \cite{gaussianbev} tasks, we expect them to be also helpful for 3D object detection, which is yet under-explored in the community.

\begin{figure}[t!]
      \centering
        \includegraphics[width=8cm]{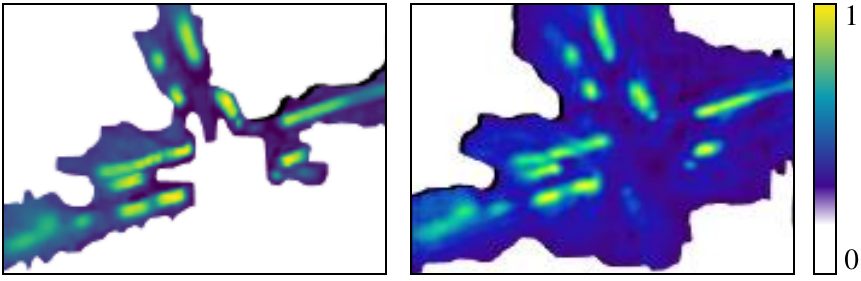}
        \caption{Comparison of BEV features before (left figure) and after (right figure) using previous radiance fields.}
        \label{previous vis}
    \end{figure}

To explore the impact of radiance fields on the detection task, we directly apply the previous radiance fields and their rendering approach to an existing detector following the above methods, but we observe a decline in detection performance (Fig. \ref{previous performance}) due to the strong responses on the background caused by rendering the whole scene (Fig. \ref{previous vis}). We argue that detection differs from the above perception tasks, in the way that it focuses on regions of interest instead of the entire scene, making the scene-centric rendering approach unsuitable for the detection task. 
Therefore, it is necessary to investigate how to make proper use of the radiance fields to enhance the detection performance.


In this paper, we propose a geometric feature enhancement approach based on the radiance fields to focus on modeling foreground objects of interest in both 3D voxel and 2D BEV space.
For 3D voxel feature enhancement, we employ object-centric radiance fields (OcRF) to enhance foreground features via an auxiliary task of rendering 3D foreground objects. Specifically, we use the hybrid generalizable radiance fields combining NeRF and 3DGS to enhance 3D geometry estimation, and we develop an object-centric rendering approach tailored for the detection task by narrowing down the optimization goal from scene-level to object-level. This enables the radiance fields to more effectively and efficiently focus on modeling foreground objects of interest. It is worth noting that the rendering approach as an auxiliary head for training can be removed during model inference to reduce the network's computational burden.
For 2D BEV feature enhancement, we propose height-aware opacity-based attention (HOA) to enhance foreground features based on geometric opacity, the side-product of rendering, which suggests the existence of foreground objects. Specifically, considering the variations in object distributions across different height levels (such as the presence of pedestrians and trees in lower-height levels and the absence of pedestrians in higher ones), we generate opacity-based attention maps at different height levels separately via multiple networks in parallel. These weights are applied to the BEV features, enabling the model to adaptively emphasize critical foreground features while suppressing invalid background noise.

Our contributions are summarized as follows:

\begin{itemize}
    \item To our best knowledge, we are the first to apply radiance fields to multi-view 3D object detection in autonomous driving. We employ 3D foreground object rendering based on object-centric radiance fields as an auxiliary task to enhance geometry estimation within 3D voxel space.
    
    
    \item We further make use of opacity -the side-product of rendering- to enhance the foreground BEV features while suppressing irrelevant background noises. Considering the variation of object distributions at different height levels, we generate weights based on opacity at different height levels separately via multiple networks in parallel.
    

    \item Our method is evaluated on the nuScenes $\tt{validation}$ and $\tt{test}$ datasets \cite{caesar2020nuscenes}, outperforming previous state-of-the-art methods.
    
\end{itemize}
\section{Related Work}
Since we focus on the task of multi-view 3D object detection and incorporate radiance fields to assist detection, we review recent works in multi-view 3D object detection, including depth-based and query-based detectors, and provide a brief overview of the radiance fields for perception.

\subsection{Multi-View 3D Object Detection}
\textbf{Depth-based detectors.} This type of detector estimates 3D geometry by learning depth distributions. BEVDet \cite{bevdet} follows LSS \cite{lss} to lift multi-view 2D camera features into 3D through implicit pixel-wise discrete depth prediction. BEVDepth \cite{bevdepth} emphasizes the importance of depth in 3D geometry estimation and introduces explicit depth supervision. BEV-SAN \cite{bevsan} highlights the height information during view transformation. BEVNeXt \cite{bevnext} introduces conditional random fields to improve the accuracy of depth estimation. SA-BEV \cite{sabev} introduces SA-BEVPool to enhance the geometry estimation of foreground objects by leveraging semantic segmentation as supervision. DualBEV \cite{dualbev} further extends this approach and proposes Prob-LSS to ignore invalid background features. Benefiting from the low computational cost of the depth-based methods, we choose this type of detector as our baseline, and introduce the radiance fields to enhance the geometry information of the foreground features, while emphasizing the height-aware representations in the radiance field.

\noindent \textbf{Query-based detectors.} This type of detector directly predicts 3D geometry in the BEV space by utilizing cross-attention to relate 2D image features with 3D queries. BEVFormer \cite{bevformer, bevformerv2} constructs a unified BEV space and designs spatial cross-attention to query the spatial 2D features. Based on BEVFormer, HeightFormer \cite{heightformer} introduces the height predictor to highlight the height information in BEV space. DETR3D \cite{detr3d} employs a sparse set of learnable reference points as 3D object queries to index 2D features. Based on this, Sparse4D \cite{sparse4d} proposes a deformable 4D aggregation module to boost detection performance based on multi-dimensional features. SparseBEV \cite{sparsebev} uses pillars instead of reference points as the 3D queries. QAF2D \cite{QAF2D} introduces a pre-trained 2D detector to generate precise sparse 3D queries. RayDN \cite{raydn} samples along camera rays to construct hard negative queries. OPEN \cite{open} employs object-wise depth instead of typical pixel-wise depth for accurate detection. Our approach is also compatible with these methods, further enhancing the detection performance. 

    \begin{figure*}[h]
      \centering
       \includegraphics[width= \linewidth]{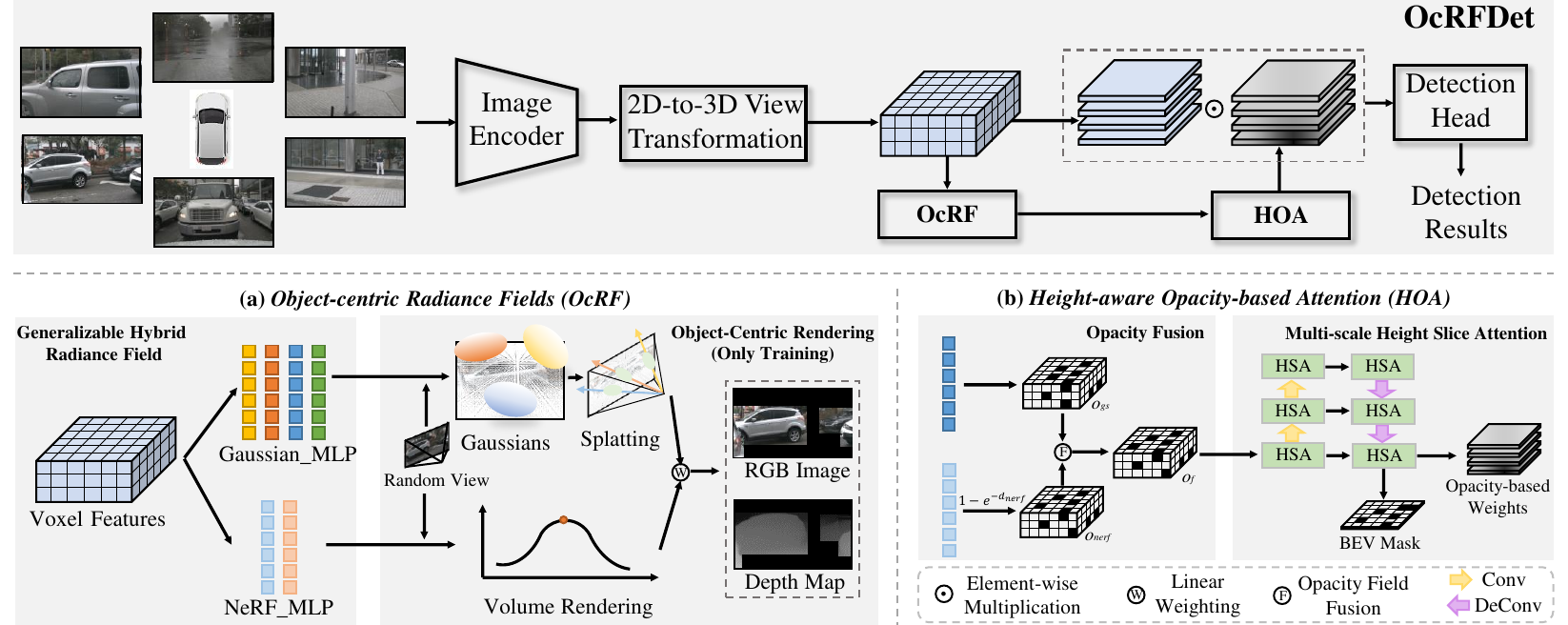}
       \caption{Overview of our method. It introduces the object-centric radiance fields (Sec. \ref{OcRF}) and height-aware opacity-based attention (Sec. \ref{HOA}) within an existing detector to enhance 3D geometry estimation in both 3D voxel and 2D BEV space.}
      \label{pipeline}
    \end{figure*}  
    
\subsection{Radiance Field in Perception}
Recently, radiance field methods have demonstrated strong geometry estimation capabilities in perception tasks such as occupancy prediction and BEV segmentation. In occupancy prediction, UniOcc \cite{uniocc} and RenderOcc \cite{renderocc} explicitly impose spatial geometric constraints through volumetric ray rendering. Based on this, HybridOcc \cite{hybridocc} introduces 3D occupancy-aware ray sampling, allowing the model to focus on occupied voxels. GaussianFormer \cite{gaussianformer} uses 3D Gaussian representations to model 3D scenes and performs occupancy prediction through Gaussian-to-voxel splatting. In BEV segmentation, GaussianBEV \cite{gaussianbev} uses 3D Gaussian representations for scene modeling and performs BEV segmentation via BEV splatting. The aforementioned scene-level radiance field methods are unsuitable for the detection task due to the introduction of irrelevant background information. To address this, we propose a novel approach tailored for the detection task.

\section{Methodology}
In this section, we first give an overview of our proposed multi-view 3D object detector, and then provide a detailed introduction to our proposed geometric enhancement method based on radiance fields.

\subsection{Overview}
We propose OcRFDet, a geometric feature enhancement approach based on the radiance fields, to enhance the detection network's 3D geometric modeling capabilities, building upon an existing detector. As illustrated in Fig. \ref{pipeline}, our approach consists of two important modules: OcRF and HOA. In the following, we briefly describe the whole pipeline.

\noindent \textbf{Image encoder.} 
We use an image backbone with a feature pyramid network as the image encoder, which takes multi-view images as inputs, and encodes them into high-level image features.

\noindent \textbf{2D-to-3D view transformation.} 
This module transforms the image features from the multiple views to the bird's eye view. Following \cite{bevdepth}, we employ its Depth Net to predict depth probabilities, and a view transformer to lift the image features into pre-defined voxel space. 

\noindent \textbf{Object-centric radiance field.} 
The OcRF module is designed to enhance the voxel features. We adopt hybrid generalizable radiance fields to improve generalization ability and 3D geometry estimation, and design an object-centric rendering approach tailored for detection task. Please refer to Sec. \ref{OcRF} for more details.

\noindent \textbf{Height-aware opacity-based attention.} 
The HOA module is designed to enhance the BEV feature. We introduce Height Slice Attention (HSA) into multi-scale opacity fields to obtain opacity-based attention maps at different height levels, and these weights are applied to the BEV features across different channel levels for enhancement. Please refer to Sec. \ref{HOA} for more details.

\noindent \textbf{Detection head.} 
Based on the refined BEV features, we adopt a center-based 3D object detection head \cite{centerpoint} to predict the object categories and 3D bounding boxes.

\noindent \textbf{Optimization.}
We jointly optimize the detector and the radiance field branch end-to-end. The entire loss function is formulated as a summation of the detection loss, the rendering loss, and the BEV mask loss:
    \begin{equation}
       L = L_{det} + L_{render} + L_{mask},
    \end{equation}
\noindent where the detection loss $L_{det}$ is the same as the baseline detector. Please refer to Sec. \ref{rendering loss} for rendering loss $L_{render}$ and Sec. \ref{HSA} for BEV mask loss $L_{mask}$.

\subsection{Object-Centric Radiance Field} \label{OcRF}
As shown in Fig. \ref{pipeline} (a), the object-centric radiance fields consist of hybrid generalizable radiance fields and an object-centric rendering approach. In the following, we provide a detailed explanation of each component.

\subsubsection{Hybrid Generalizable Radiance Fields}
Since traditional radiance fields rely on per-scene optimization, it is crucial to develop representations that can generalize across unseen scenes, particularly for the detection task. To this end, following \cite{generalizable3dgs} and \cite{nerfdet}, we employ multiple MLPs to decode 3D voxel features into randiance field representations, rather than typically encoding them from 2D RGB images. Moreover, we adopt a hybrid representation from 3DGS and NeRF to enhance generalization ability and improve geometric estimation.

Specifically, for the 3DGS representations, the attributes include position $p_{gs}$, scale $s_{gs}$, rotation $r_{gs}$, opacity $o_{gs}$, and color $c_{gs}$. For position, we directly scale the voxel coordinates to align with the actual perception range. Meanwhile, the remaining attributes are predicted through a Gaussian$\_$MLP. The specific predictions are formulated as:
    \begin{equation}
        \begin{aligned}
        s_{gs} &= \text{Softplus}(S \text{\_} MLP(F_v)), \\ 
        r_{gs} &= \text{Normalize}(R \text{\_} MLP(F_v)), \\ 
        o_{gs} &= \text{Sigmoid}(O \text{\_} MLP(F_v)), \\ 
        c_{gs} &= \text{Sigmoid}(C \text{\_} MLP(F_v)),
        \end{aligned}
    \end{equation}
\noindent where $S \text{\_} MLP(\cdot)$, $R \text{\_} MLP(\cdot)$, $O \text{\_} MLP(\cdot)$, and $C \text{\_} MLP(\cdot)$ each consists of a two-layer MLP. Besides, following \cite{generalizable3dgs}, we apply specific activation functions to ensure the predicted values align with their physical ranges. For the NeRF representations, the attributes include density $d_{nerf}$ and color weight $w_{nerf}$. Similarly, these attributes are predicted through a NeRF$\_$MLP, and the network corresponding to each attribute consists of a two-layer MLP, and we apply specific activation functions to ensure the reasonability of predicted values.

As a result, we obtain the necessary attribute values, which are then used for rendering. These attribute values, predicted from high-level 3D voxel features, enable our model to generalize across diverse scenes.

\subsubsection{Object-Centric Rendering}
With the radiance field attributes available, we apply corresponding rendering techniques to generate RGB images and depth maps from specific viewpoints. Here, we randomly select one viewpoint from multiple views to render for a low computational cost. Specifically, for 3DGS attributes $\left\{p_{gs}, s_{gs}, r_{gs}, o_{gs}, c_{gs}\right\}$, we apply a degenerate depth-guided splatting regularization method \cite{depth3dgs}, establishing a one-to-one correspondence between each 3DGS and a pixel to render the selected viewpoint. For the NeRF attributes $\left\{d_{nerf}, w_{nerf}\right\}$, we compute the opacity attribute $o_{nerf}$, as follows: $o_{nerf}=1-e^{-d_{nerf}}$, and the color attribute $c_{nerf}$ is then obtained by applying the predicted color weights to the image plane features. After computing these NeRF attributes, we perform simple degenerate volume rendering, where only one point is sampled per ray to render the selected viewpoint, while producing the corresponding depth as the distance between the sampled point and the camera. 

Unlike previous methods \cite{renderocc,gaussianbev} that optimize the entire scene, we only compute the loss values for the foreground regions based on the 2D labels, mapped from the 3D boxes to the image plane of the corresponding viewpoint. This object-centric rendering encourages both the Gaussian$\_$MLP and NeRF$\_$MLP to focus on modeling the foreground objects, thereby enhancing the quality of the voxel features. To maintain geometric consistency between the two radiance field networks, we further optimize the linear fusion of their rendered results:
    \begin{equation}
        \begin{aligned}
            I_{f} &= \alpha \cdot I_{gs} + \beta \cdot I_{nerf}, \\
            D_{f} &= \alpha \cdot D_{gs} + \beta \cdot D_{nerf},
        \end{aligned}
        \label{fusion}
    \end{equation}
\noindent where $\left\{ I_{gs}, I_{nerf} \right\}$ and $\left\{ D_{gs}, D_{nerf} \right\}$ are the RGB images and depth maps rendered by 3DGS and NeRF, respectively; $\alpha$ and $\beta$ are the learnable hyper-parameters, and $\alpha + \beta = 1 \; ( \alpha, \beta > 0 )$. Additionally, the scene-level optimization is employed as a warm-up phase to guide the initial training of the Gaussian$\_$MLP and NeRF$\_$MLP, and the whole rendering process is not involved during inference.

\subsubsection{Rendering Loss} \label{rendering loss}
We optimize the radiance field branch with the mean squared error loss,  structural similarity index measure loss \cite{ssim}, and $l1$ loss, as follows:
    \begin{equation}
       L_{render}= \lambda_{mse} \cdot L_{mse} + \lambda_{ssim} \cdot L_{ssim} + \lambda_{l1} \cdot L_{l1},
    \end{equation}
\noindent where $L_{mse}$ and $L_{ssim}$ are typically used to render the RGB images, and $L_{l1}$ is additionally introduced to enhance geometric modeling for depth map rendering; $\lambda_{mse}$,  $\lambda_{ssim}$ and $\lambda_{l1}$ denote the weights for balancing loss items. Specifically, $L_{mse}$ and $L_{ssim}$ can be calculated as:
    \begin{equation}
       L_{mse}= 
            \frac{1}{N} \sum_{i=1}^{N} \sum_{\hat{I} \in I_p} [(I_i - \hat{I}_i) \odot M_i]^2,
    \end{equation}
    \begin{equation}
       L_{ssim}= 
           \frac{1}{N} \sum_{i=1}^{N} \sum_{\hat{I} \in I_p} 1 - SSIM(I_i \odot M_i, \hat{I}_i \odot M_i),
    \end{equation}
\noindent where $N$ is the number of samples; $I$ represents the ground truth of RGB images; $M$ is the 2D binary mask, where pixels inside the 2D ground truth boxes are set to 1 and those outside are set to 0; $I_p$ contains the predicted results $\left\{ I_{gs}, I_{nerf}, I_{f} \right\}$; $SSIM(\cdot)$ represents the SSIM function. Similarly, $L_{1}$ can be calculated as:
    \begin{equation}
       L_{l1}= 
            \frac{1}{N} \sum_{i=1}^{N} \sum_{\hat{D} \in D_p} |(D_i - \hat{D}_i) \odot M_i|,
    \end{equation}
\noindent where $D$ represents the ground truth of depth maps containing the cropped foreground objects; $D_p$ contains the predicted results $\left\{ D_{gs}, D_{nerf}, D_{f} \right\}$.

\subsection{Height-Aware Opacity-based Attention} \label{HOA}
As shown in Fig. \ref{pipeline} (b), the HOA consists of an opacity fusion and multi-scale height slice attention. In the following, we provide a detailed explanation of each component.

\subsubsection{Opacity Fusion}
To enhance the geometric information of BEV features, we leverage the opacity field, a side-product of radiance field rendering, as it inherently encodes geometric knowledge within the radiance fields. We introduce an opacity fusion that fuses two geometric opacity fields $\left\{o_{gs}, o_{nerf}\right\}$ through a cross-attention mechanism, establishing fine-grained associations between the two fields and enhances the representation of geometric information:
    \begin{equation}
        o_{f} = CA(Q, K, V), 
    \end{equation}
    \small
    \vspace{-0.8cm}
    \begin{align} 
            \begin{cases} Q = o_{nerf}, K = o_{gs}, V = o_{gs}, \; \text{if} \; \alpha \leq \beta \\
            Q = o_{gs}, K = o_{nerf}, V = o_{nerf}, \; \text{if} \; \alpha > \beta 
            \end{cases},
    \end{align}
    \normalsize
where \noindent $CA(\cdot)$ denotes the cross-attention operation; $\alpha$ and $\beta$ are the learnable hyper-parameters in Eq. \ref{fusion}. When the rendered images from some certain radiance field (3DGS or NeRF) have a higher weight ($\alpha$ or $\beta$), it reflects its superior rendering quality and high-quality geometry opacity. In such a case, the corresponding opacity field serves as the query $Q$ in the cross-attention mechanism, while the other field acts as the key $K$ and value $V$. 

\subsubsection{Multi-Scale Height Slice Attention} \label{HSA}
Based on the above fused opacity field, we further generate an attention map, which will be applied to the BEV features so as to highlight foreground objects.
    \begin{figure}[t!]
      \centering
       \includegraphics[width=8.0cm]{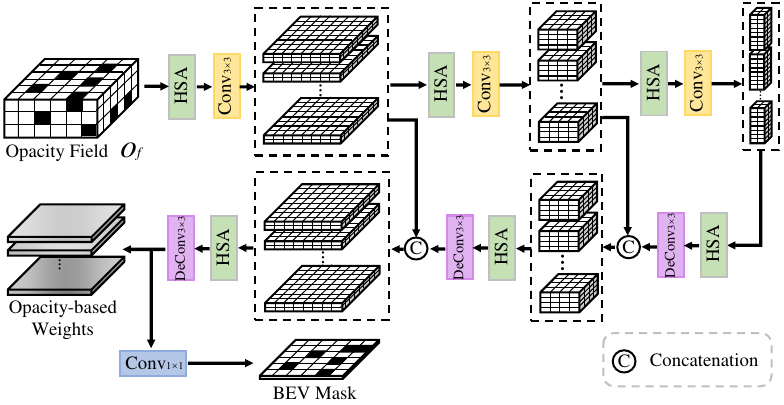}
       \caption{Illustration of multi-scale height slice attention.}
      \label{Fig. MHSA}
    \end{figure}
    
\noindent \textbf{Height slice attention.} 
To refine the opacity field based on inner dependencies, we use self-attention at each layer. Inspired by previous works \cite{heightformer,bevsan} that highlight the importance of height information in BEV features, and consider the diverse distributions of objects at different heights, we employ multiple self-attention blocks in parallel, each for a certain height range. Specifically, the given opacity field is divided into $k$ groups along height. Each of them is separately processed through MaxPooling, followed by a 1 $\times$ 1 convolution, converting the local geometric opacity information for each height interval into a corresponding geometric attention map. These attention maps are then concatenated in the order of height. Finally, a sigmoid activation function projects the attention map values to a range between $0$ and $1$, aligning them with their physical ranges.

\noindent \textbf{Multi-scale integration.} 
As shown in Fig. \ref{Fig. MHSA}, a three-layer opacity pyramid is employed to integrate opacity information across multiple scales. Specifically, 3 $\times$ 3 convolution and deconvolution operations are applied to capture opacity at different scales. During scale transformation, HSA highlights the essential opacity information along the height dimension, while cascade connections retain original opacity details, enhancing the integration of multi-scale opacity. 

\noindent \textbf{BEV mask prediction.}
To suppress noises from the background regions, we add an auxiliary task of BEV mask prediction, following \cite{seabird}. This mask is predicted by a 1 $\times$ 1 convolution and optimized by a combination of binary cross-entropy loss and dice loss \cite{diceloss}, as follows:
    \begin{equation}
       L_{mask}= \lambda_{bce} \cdot L_{bce} + \lambda_{dice} \cdot L_{dice},
    \end{equation}
\noindent where $\lambda_{bce}$ and $\lambda_{dice}$ denote the weights for balancing loss items.
\section{Experiments}
In this section, we will first introduce the dataset and evaluation metrics, followed by the implementation details. Then, we compare our method with the state-of-the-art methods. Finally, we will show the ablation studies and visualization.

    \begin{table*}[th!]
        \centering
        \caption{Comparisons with the state of the art on the nuScenes $\tt{validation}$ set. All methods use ResNet-50 \cite{he2016resnet} with FPN \cite{lin2017fpn} as the image encoder, and the image resolution is set to 256 $\times$ 704.}
        \setlength{\tabcolsep}{2.3mm}{
        \scalebox{0.86}{
        \begin{tabular}{>{\centering\arraybackslash}m{2.7cm} | c | c c |c c c c c c} 
        \hline
        Methods & Frames & mAP $\uparrow$ & NDS $\uparrow$ & mATE $\downarrow$ & mASE $\downarrow$ & mAOE $\downarrow$ & mAVE $\downarrow$ & mAAE $\downarrow$ \\
        \hline
        BEVDet \cite{bevdet} & 1 & 29.8 & 37.9 & 0.725 & 0.279 & 0.589 & 0.860 & 0.245 \\
        BEVFormer \cite{bevformer} & 1 & 29.7 & 37.9 & 0.739 & 0.281 & 0.601 & 0.833 & 0.242 \\
        FB-BEV \cite{fbbev} & 1 & 31.2 & 40.6 & 0.702 & 0.275 & 0.518 & 0.777 & 0.227 \\
        BEVDepth \cite{bevdepth} & 1 & 34.2 & 40.7 & 0.645 & 0.273 & 0.599 & 0.890 & 0.240 \\
        DualBEV \cite{dualbev} & 1 & 35.2 & 42.5 & 0.640 & 0.271 & 0.542 & 0.838 & 0.216 \\
        \rowcolor{gray!20}
        \textbf{OcRFDet (Ours)} & 1 & \textbf{36.8} & \textbf{43.4} & 0.620 & 0.269 & 0.519 & 0.848 & 0.242 \\
        \hline
        BEVDet4D \cite{bevdet4d} & 2 & 31.6 & 44.9 & 0.691 & 0.281 & 0.549 & 0.378 & 0.195 \\
        BEVFormer \cite{bevformer} & 2 & 33.0 & 45.9 & 0.686 & 0.272 & 0.482 & 0.417 & 0.201 \\
        BEVDepth \cite{bevdepth} & 2 & 36.8 & 48.5 & 0.609 & 0.273 & 0.507 & 0.406 & 0.196 \\
        BEVStereo \cite{bevstereo} & 2 & 37.2 & 50.0 & 0.598 & 0.270 & 0.438 & 0.367 & 0.190 \\
        FB-BEV \cite{fbbev} & 2 & 37.8 & 49.8 & 0.620 & 0.273 & 0.444 & 0.374 & 0.200 \\
        SA-BEV \cite{sabev} & 2 & 37.8 & 49.9 & 0.617 & 0.270 & 0.441 & 0.370 & 0.206 \\
        DualBEV \cite{dualbev} & 2 & 38.0 & 50.4 & 0.612 & 0.259 & 0.403 & 0.370 & 0.207 \\
        \rowcolor{gray!20}
        \textbf{OcRFDet (Ours)} & 2 & \textbf{40.0} & \textbf{50.9} & 0.582 & 0.277 & 0.518 & 0.375 & 0.216 \\
        \hline
        \end{tabular}
            }
        }
        \label{nuScenes validation}
    \end{table*}

    \begin{table*}[th!]
        \centering
        \caption{Comparisons with the state of the art on the nuScenes $\tt{test}$ set. By default, the image backbone is set as V2-99 \cite{v299}, initialized from DD3D \cite{dd3d}; the image resolution is set to 640 $\times$ 1600. $\dag$ indicate the use of ConvNeXt-B \cite{convnext} as the image backbone. All methods consider the historical temporal information of 8 frames and employ 60-epoch training schemes for fair comparison.}
        \setlength{\tabcolsep}{2.7mm}{
        \scalebox{0.86}{
        \begin{tabular}{>{\centering\arraybackslash}m{3.2cm} | c c |c c c c c c} 
        \hline
        Methods & mAP $\uparrow$ & NDS $\uparrow$ & mATE $\downarrow$ & mASE $\downarrow$ & mAOE $\downarrow$ & mAVE $\downarrow$ & mAAE $\downarrow$ \\
        \hline
        SOLOFusion $\dag$ \cite{solofusion} & 54.0 & 61.9 & 0.453 & 0.257 & 0.376 & 0.276 & 0.148 \\
        SparseBEV \cite{sparsebev} & 54.3 & 62.7 & 0.502 & 0.244 & 0.324 & 0.251 & 0.126 \\
        StreamPETR \cite{streampetr} & 55.0 & 63.6 & 0.479 & 0.239 & 0.317 & 0.241 & 0.119 \\
        DualBEV \cite{dualbev} & 55.2 & 63.4 & 0.414 & 0.245 & 0.377 & 0.252 & 0.129 \\
        VideoBEV $\dag$ \cite{videobev} & 55.4 & 62.9 & 0.457 & 0.249 & 0.381 & 0.266 & 0.132 \\
        Sparse4Dv2 \cite{sparse4d} & 55.6 & 63.8 & 0.462 & 0.238 & 0.328 & 0.264 & 0.115 \\
        BEVNeXt \cite{bevnext} & 55.7 & 64.2 & 0.409 & 0.241 & 0.352 & 0.233 & 0.129 \\
        RecurrentBEV \cite{recurrentbev} & 56.4 & 64.4 & 0.427 & 0.238 & 0.327 & 0.257 & 0.133 \\
        QAF2D \cite{QAF2D} & 56.6 & 64.2 & 0.461 & 0.240 & 0.326 & 0.261 & 0.121 \\
        RayDN \cite{raydn} & 56.5 & 64.5 & 0.461 & 0.241 & 0.322 & 0.239 & 0.114 \\
        OPEN \cite{open} & 56.7 & 64.4 & 0.456 & 0.244 & 0.325 & 0.240 & 0.129 \\
        \rowcolor{gray!20}
        \textbf{OcRFDet (Ours)} & \textbf{57.2} & \textbf{64.8} & 0.421 & 0.253 & 0.331 & 0.248 & 0.130 \\
        \hline
        \end{tabular}
            }
        }
        \label{nuScenes test}
    \end{table*}

\subsection{Experimental Setup}
\noindent \textbf{Datasets and evaluation metrics.} 
We evaluate our method on the nuScenes $\tt{validation}$ and $\tt{test}$ datasets \cite{caesar2020nuscenes}.
We use the mean Average Precision (mAP) and nuScenes detection scores (NDS) to evaluate our detection results, where NDS is a comprehensive metric that combines the errors of object translation (mATE), scale (mASE), orientation (mAOE), velocity (mAVE), and attribute (mAAE).

\noindent \textbf{Implementation details.} 
We select DualBEV \cite{dualbev} as our baseline. The BEV grid size is set to 128 $\times$ 128 for the nuScenes $\tt{validation}$ dataset, and adjusted to 256 $\times$ 256 for the nuScenes $\tt{test}$ dataset. We integrate previous multiple frames following BEVDet4d \cite{bevdet4d}. Following 4DGS \cite{4DGS}, the scene-level optimization is used as a warm-up during the first 2 epochs here. The hyper-parameter $k$ in HSA is set to 4. We set the loss weights $\lambda_{mse}$, $\lambda_{ssim}$, $\lambda_{l1}$, $\lambda_{bce}$, $\lambda_{dice}$ as 10, 1, 1, 10, and 10 to balance the loss values of auxiliary tasks and the detection task. Please refer to the supplementary material for more training details.
    
\subsection{Main Results}
We compare our OcRFDet with previous state-of-the-art methods on the nuScenes $\tt{validation}$ and $\tt{test}$ datasets. As illustrated in Tab. \ref{nuScenes validation}, for the nuScenes $\tt{validation}$ dataset, our OcRFDet achieves 36.8$\%$ mAP and 43.4$\%$ NDS performance with one single frame, which outperforms the state-of-the-art method (DualBEV) by 1.6 pp w.r.t. mAP and 0.9 pp w.r.t. NDS. When using 2 consecutive frames as inputs, OcRFDet achieves 40.0$\%$ mAP and 50.4$\%$ NDS performance, which outperforms the DualBEV by 2.0 pp w.r.t. mAP and 0.5 pp w.r.t. NDS. These results demonstrate the effectiveness of our method. 

Furthermore, as illustrated in Tab. \ref{nuScenes test}, for the nuScenes $\tt{test}$ dataset, our OcRFDet achieves 57.2$\%$ mAP and 64.8$\%$ NDS performance when using a longer 8-frame sequence as inputs, which outperforms DualBEV by 2.0 pp w.r.t. mAP and 1.1 pp w.r.t. NDS. Additionally, when compared to the state-of-the-art method (OPEN), which focuses on the temporal module, our OcRFDet maintains superiority and yields a new state-of-the-art result. These experiments further demonstrate the effectiveness of our method. More comparisons on the Waymo Open dataset \cite{Waymo} are provided in the supplemental material.

\subsection{Ablation Studies}
In this section, we conduct ablation studies to investigate the effectiveness of our OcRFDet. All method variants with one frame are trained with 20-epoch training strategy on the nuScenes $\tt{validation}$ dataset. The latency tests are conducted on a single RTX 3090 GPU without TensorRT.

    \begin{figure}[t]
        \centering
        \includegraphics[width=8.1cm]{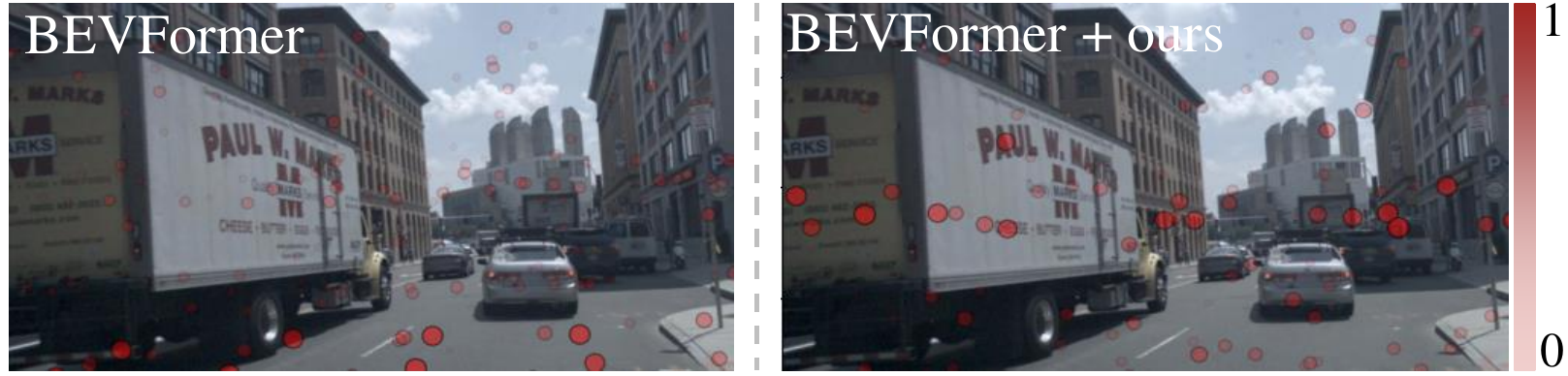}
        \caption{Visualization results of the 2D reference points.
        } \label{reference points}
    \end{figure}
    \begin{figure}[t]
        \centering
        \captionof{table}{Comparisons of BEVFormer-tiny \cite{bevformer} as the baseline.}
        \centering
        \setlength{\tabcolsep}{0.8mm}{
        \scalebox{0.90}{
        \begin{tabular}{c | c c |c c } 
        \hline
        Methods & Backbone & BEV size & mAP $\uparrow$ & NDS $\uparrow$ \\
        \hline
        BEVFormer & \; \multirow{2}{1.7cm}{ResNet-50} & \multirow{2}{1.0cm}{50$\times$50} & 25.2 & 35.5 \\
        \textbf{BEVFormer + Ours} & & & \textbf{27.4} & \textbf{36.9} \\
        \hline
        \end{tabular}
            }
        }
        \label{bevformer}
    \end{figure}
    \begin{table}[t]
        \caption{Ablations of each component.}
        \centering
        \setlength{\tabcolsep}{2.4mm}{
        \scalebox{0.90}{
        \begin{tabular}{ >{\centering\arraybackslash}m{2.5cm} | c c | >{\centering\arraybackslash}m{2.5cm}} 
        \hline
         Methods & mAP $\uparrow$ & NDS $\uparrow$ & Latency (ms) $\downarrow$ \\
        \hline
        Baseline & 34.1 & 40.3 & \textbf{93.46} \\
        + OcRF & 35.0 & 41.9 & 109.89 \\
        + HOA & \textbf{35.5} & \textbf{42.2} & 112.36 \\
        \hline
        \end{tabular}
            }
        }
        \label{each component}
    \end{table}

\noindent \textbf{Generalization to query-based baseline.} 
As shown in Tab. \ref{bevformer}, when using BEVFormer-tiny \cite{bevformer} as our baseline on the nuScenes $\tt{validation}$ set, our method brings the improvements of 2.2 pp w.r.t. mAP and 1.4 pp w.r.t. NDS. Furthermore, as shown in Fig. \ref{reference points}, we visualize the 2D reference points with their weights in the BEVFormer encoder, and clearly find that our method makes the reference points of foreground have higher weights. These results demonstrate our method generalizes well to query-based detectors. 

\noindent \textbf{Effectiveness of each component.}
As shown in Tab. \ref{each component}, compared with the baseline, our OcRF brings 0.9 pp w.r.t. mAP and 1.6 pp w.r.t. NDS improvements by enhancing 3D geometric features. On top of it, further adopting our HOA improves the performance by 0.5 pp w.r.t. mAP and 0.3 pp w.r.t. NDS by enhancing 2D BEV features. When combining both components, the results are improved significantly by 1.4 pp w.r.t. mAP and 1.9 pp w.r.t. NDS. These results demonstrate the effectiveness of each component. Additionally, we also test the latency of each component. The two components reveal an extra latency of 18.90 ms in total, with 16.43 ms from OcRF and 2.47 ms from HOA, which is minimal and thus friendly to applications.
    
\noindent \textbf{Ablations of hybrid radiance fields.}
As shown in Tab. \ref{hybrid radiance field}, using a single radiance field leads to a drop in detection performance and rendering quality. These suggest a strong correlation between geometric modeling capacity and detection accuracy. Hybrid radiance fields improve geometric consistency and structural completeness, enhancing feature quality and ultimately boosting detection performance.

\noindent \textbf{Effectiveness of depth rendering.}
As shown in Tab. \ref{depth rendering}, compared with only rendering RGB images, the depth rendering brings 0.4 pp w.r.t. mAP and 0.8 w.r.t. NDS performance improvements by introducing explicit 3D geometric information. This result demonstrates the effectiveness of depth rendering in enhancing detection performance.

    \begin{table}[t]
        \caption{Ablations of hybrid radiance fields.}
        \centering
        \setlength{\tabcolsep}{2.4mm}{
        \scalebox{0.90}{
        \begin{tabular}{ >{\centering\arraybackslash}m{1.0cm} >{\centering\arraybackslash}m{1.0cm} | >{\centering\arraybackslash}m{1.0cm} >{\centering\arraybackslash}m{1.0cm} | >{\centering\arraybackslash}m{2.0cm}} 
        \hline
        3DGS & NeRF & mAP $\uparrow$ & NDS $\uparrow$ & SSIM $\uparrow$ \\
        \hline
        \ding{51} & & 34.7 & 41.4 & 0.6803 \\
         & \ding{51} & 34.5 & 41.3 & 0.6678\\
        \ding{51} & \ding{51} & \textbf{35.0} & \textbf{41.9} & \textbf{0.7073} \\
        \hline
        \end{tabular}
            }
        }
        \label{hybrid radiance field}
    \end{table}
    \begin{table}[t]
        \caption{Ablations of depth rendering.}
        \centering
        \setlength{\tabcolsep}{2.4mm}{
        \scalebox{0.90}{
        \begin{tabular}{ >{\centering\arraybackslash}m{1.9cm} >{\centering\arraybackslash}m{1.9cm} | >{\centering\arraybackslash}m{1.4cm} >{\centering\arraybackslash}m{1.4cm} } 
        \hline
        RGB Image & Depth Map & mAP $\uparrow$ & NDS $\uparrow$ \\
        \hline
        & & 34.1 & 40.3 \\
        \ding{51} & & 34.6 & 41.1 \\
        \ding{51} & \ding{51} & \textbf{35.0} & \textbf{41.9} \\
        \hline
        \end{tabular}
            }
        }
        \label{depth rendering}
    \end{table}


    \begin{table}[t]
        \caption{Ablations of optimization goal.}
        \centering
        \setlength{\tabcolsep}{2.2mm}{
        \scalebox{0.77}{
        \begin{tabular}{c | c c | c c} 
        \hline
        \multirow{2}{2.8cm}{Optimization Goal} & \multirow{2}{1.0cm}{mAP $\uparrow$} & \multirow{2}{1.0cm}{NDS $\uparrow$} & \multicolumn{2}{c}{SSIM $\uparrow$} \\
        \cline{4-5}
        & & & Foreground & Entire scene \\
        \hline
        scene & 33.8 & 40.1 & 0.6211 & 0.5731 \\
        object & 34.7 & 41.5 & 0.6793 & 0.4825 \\
        scene + object & \textbf{35.0} & \textbf{41.9} & \textbf{0.7073} & \textbf{0.6134} \\
        \hline
        \end{tabular}
            }
        }
        \label{rendering optimization goal}
    \end{table}

\noindent \textbf{Effectiveness of rendering optimization goal.}
As shown in Tab. \ref{rendering optimization goal}, optimizing the entire scene alone results in a 0.3$\%$ mAP and 0.2$\%$ NDS performance drop compared to the baseline. When optimizing the foreground objects, we observe significant performance improvements of 0.9 pp w.r.t. mAP, 1.4 pp w.r.t. NDS, and a rendering quality improvement of 5.82 pp w.r.t. SSIM. These results indicate our object-centric rendering is effective and more suitable for the detection task. Furthermore, using scene-level optimization as a warm-up phase further improves detection performance by 0.3 pp w.r.t. mAP and 0.4 pp w.r.t. NDS, indicating that this combination offers optimal results.

    \begin{figure*}[t]
        \centering
        \includegraphics[width=17.4cm]{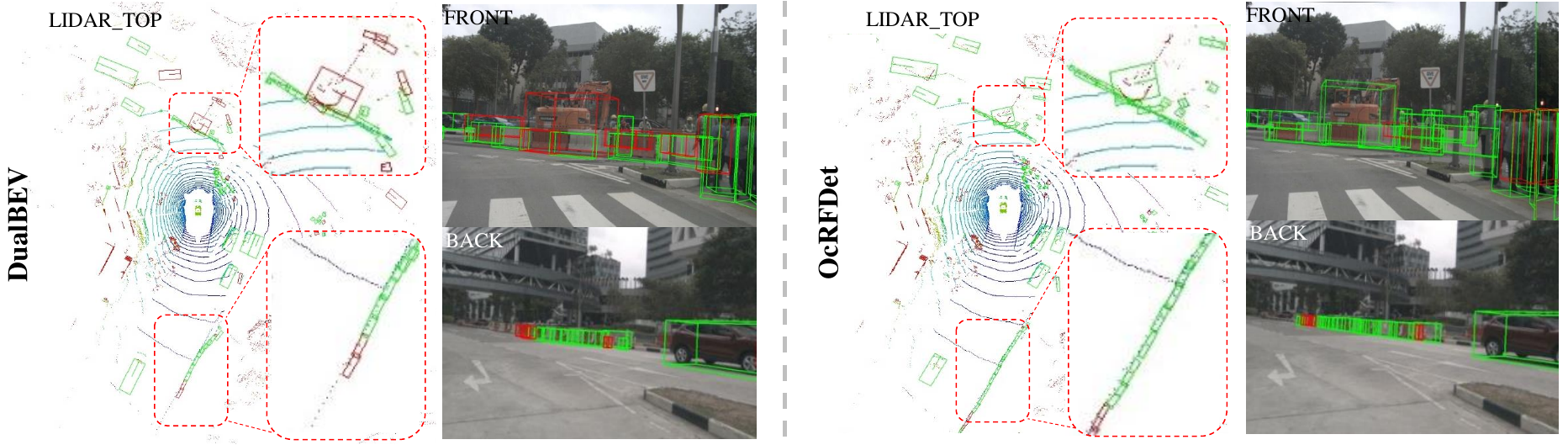}
        \caption{Qualitative detection results on images and the BEV space on the nuScenes $\tt{validation}$ set. We show the true positive boxes in \textcolor{green}{green}, and the incorrect prediction boxes in \textcolor{red}{red}. We use \textcolor{red}{red} rectangles to highlight the comparisons of ours with DualBEV.} \label{detections}
    \end{figure*}
    
\noindent \textbf{Ablations of opacity fusion.}
As shown in Tab. \ref{opacity fusion}, when only using the opacity field from 3DGS to generate opacity-based attention, the mAP drops by 0.8 pp and the NDS by 0.8 pp; when only using it from NeRF, the mAP also drops by 0.9 pp and the NDS by 1.0 pp. We attribute this to the fact that the fused opacity incorporates richer geometric information compared to a single-source opacity, thereby providing greater benefits for detection. This experiment highlights the necessity of opacity fusion. Please refer to supplemental material for the ablations of fusion strategies.

\noindent \textbf{Effectiveness of height slice attention.}
The results are shown in \ref{height slice attention}, when we remove our HSA, while retaining the multi-scale integration, the detection results drop w.r.t. both metrics. When applying a weight-sharing version of HSA by setting $k$ to 1, the detection performance also drops, which shows the rationality of our HSA design. Moreover, while more height division (larger $k$) can capture more rich height information, we find that setting $k=4$ obtains the optimal balance between performance and latency. These results demonstrate the effectiveness and rationality of our HSA module. Please refer to supplemental material for the ablations of multi-scale height slice attention.

    \begin{table}[t]
        \caption{Ablations of opacity fusion.}
        \centering
        \setlength{\tabcolsep}{2.4mm}{
        \renewcommand{\arraystretch}{1.1}
        \scalebox{0.9}{
        \begin{tabular}{>{\centering\arraybackslash}m{3.0cm} | >{\centering\arraybackslash}m{1.4cm} | >{\centering\arraybackslash}m{1.1cm} >{\centering\arraybackslash}m{1.1cm}} 
        \hline
        Opacity Source & Fusion & mAP $\uparrow$ & NDS $\uparrow$ \\
        \hline
        3DGS & \ding{55}  & 34.7 & 41.4 \\
        NeRF & \ding{55}  & 34.6 & 41.2 \\
        3DGS + NeRF & \ding{51} & \textbf{35.5} & \textbf{42.2} \\
        \hline
        \end{tabular}
            }
        }
        \label{opacity fusion}
    \end{table}
    \begin{table}[t]
        \caption{Ablations of height slice attention.}
        \centering
        \setlength{\tabcolsep}{2.4mm}{
        \scalebox{0.85}{
        \begin{tabular}{ >{\centering\arraybackslash}m{1.9cm} | >{\centering\arraybackslash}m{0.9cm} | c c | c } 
        \hline
        Strategy & $k$ & mAP $\uparrow$ & NDS $\uparrow$ & Latency (ms) \\
        \hline
        \textit{None} & - & 34.9 & 41.4 & - \\
        \hline
        \multirow{4}{0.8cm}{HSA} & 1 & 35.0 & 41.7 & 0.37 \\
        & 2 & 35.3 & 41.9 & 0.71 \\
        & 4 & \textbf{35.5} & 42.2 & 1.31 \\
        & 6 & \textbf{35.5} & \textbf{42.3} & 2.04 \\
        \hline
        \end{tabular}
            }
        }
        \label{height slice attention}
    \end{table}
    
\subsection{Visualization}
\textbf{Detection results.}
As shown in Fig. \ref{detections}, we clearly find that our method shows fewer incorrect prediction boxes, including false and missed detections, especially for occluded and distant objects. These demonstrate that our approach is effective for detecting these challenging objects. More qualitative results are provided in the supplemental material.

\begin{figure}[t!]
        \centering
        \includegraphics[width=8.1cm]{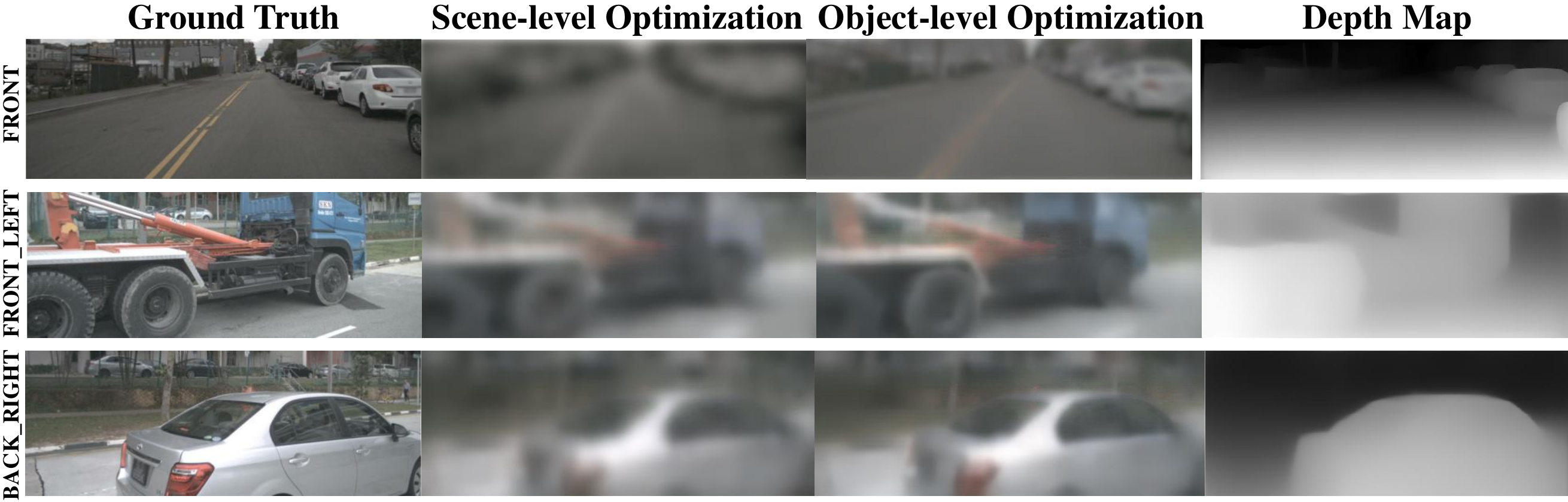}
        \caption{Visualization of rendering results.
        } \label{rendering results}
    \end{figure}

\noindent \textbf{Rendering results.}
As shown in Fig. \ref{rendering results}, we present rendering results from three different views. We observe that our object-centric optimization produces clearer foreground objects compared to scene-level optimization, with high-quality rendered depth maps. These results demonstrate that our method's radiance field achieves a stronger geometric modeling capability for foreground objects.

    \begin{figure}[t!]
        \centering
        \includegraphics[width=8.1cm]{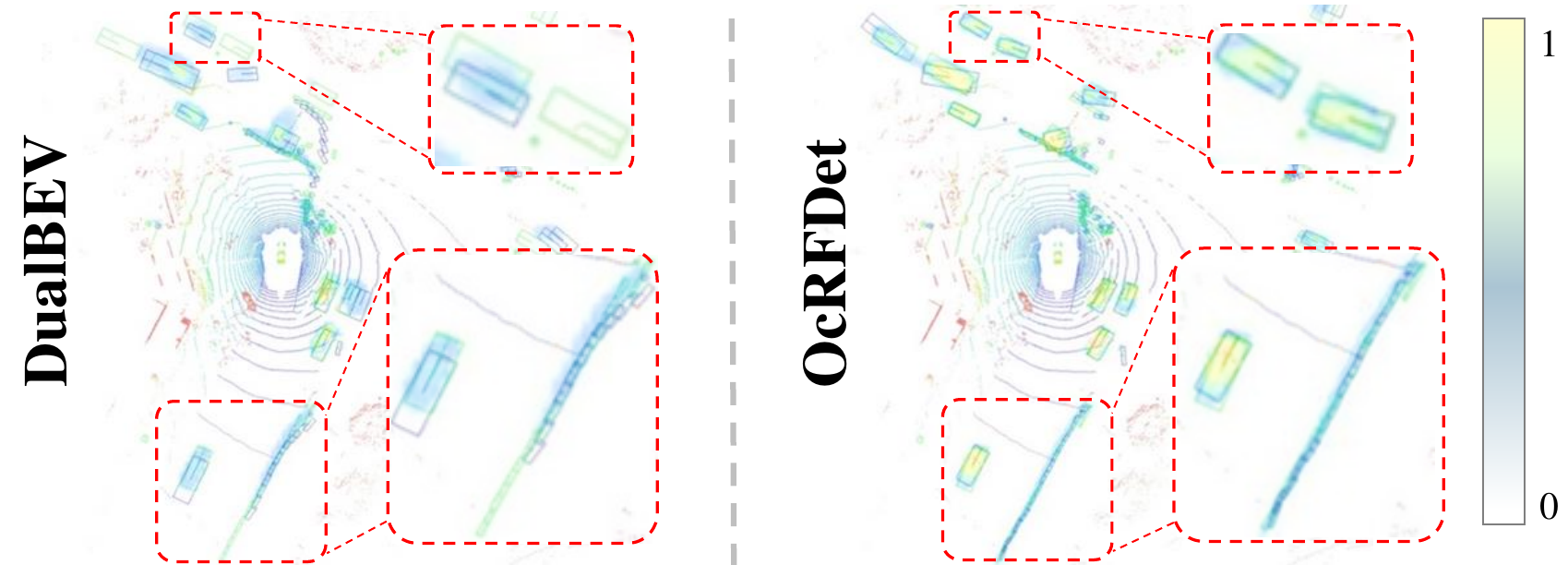}
        \caption{Comparison of BEV heatmaps. We show the ground truth boxes in \textbf{green}{green}, and the prediction boxes in \textcolor{blue}{blue}.} \label{bev feature}
    \end{figure}

\noindent \textbf{BEV heatmap comparison.}
As shown in Fig. \ref{bev feature}, our method using opacity-based attention maps exhibits higher-recall feature responses for distant objects compared to DualBEV, and the localization of feature responses is more aligned with the ground truth in crowded scenes. These indicate that our method effectively enhances features from a geometric perspective, further demonstrating the effectiveness of our approach.
\section{Conclusion}
In this paper, we propose OcRFDet, a geometric feature enhancement approach based on the radiance fields, designed to enhance the geometric information of foreground objects. Specifically, we employ object-centric radiance fields to enhance 3D voxel features via an auxiliary task of rendering foreground objects. Additionally, we propose a height-aware opacity-based attention to generate opacity-based attention maps to refine 2D BEV features. Extensive experiments demonstrate the effectiveness of our OcRFDet. 
Finally, we hope OcRFDet will further advance research on radiance fields for multi-view 3D object detection.

\maketitlesupplementary

In this document, we first provide additional implementation details of our OcRFDet (Sec. \ref{more details}). We then present extensive comparison experiments (Sec. \ref{more experiments}), ablation studies (Sec. \ref{more ablation}), and qualitative results (Sec. \ref{more qualitative}) to further validate the effectiveness of our method. Finally, we discuss the limitations of our approach and provide insights into potential future research directions (Sec. \ref{limitation}).

\section{More Implementation Details} \label{more details}
We implement the OcRFDet with PyTorch \cite{paszke2019pytorch} under the framework MMDetection3D \cite{mmdet3d2020}. On the nuScenes $\tt{validation}$ dataset, we employ a 60-epoch training scheme, and use the AdamW optimizer \cite{adamw} with the batch size set to 32 under 4 $\times$ RTX 3090 GPUs. The learning rate of the detector is initialized at 2 $\times$ $10^{-4}$ while the learning rate of radiance fields is set to 4 $\times$ $10^{-4}$ with a decay applied every one epoch. Following the baseline, we adopt the default data augmentation techniques. On the nuScenes $\tt{test}$ dataset, we employ 60-epoch training schemes with the batch size set to 8 under 8 $\times$ RTX 3090 GPUs.

\section{More Comparison Experiments} \label{more experiments}
\noindent \textbf{More results on the Waymo dataset.}
We further validate the effectiveness of our method on the Waymo Open dataset \cite{Waymo}. As shown in Tab. \ref{waymo-mini}, our method, using BEVFormer as the baseline, shows consistent improvements on Waymo-full and Waymo-mini about the LET-mAPL and LET-mAPH metrics \cite{li2023delving}. Specifically, for the Waymo-full, our method achieves 37.1\% LET-mAPL and 51.3\% LET-mAPH, which outperforms BEVFormer by 2.1 pp w.r.t. LET-mAPL and 4.2 pp w.r.t. LET-mAPH, and the state-of-the-art method VectorFormer \cite{voctorformer} by 0.3 pp w.r.t. LET-mAPL and 2.2 pp w.r.t. LET-mAPH. Moveover, for the Waymo-mini, our method also consistently achieves better results towards the baseline and the state-of-the-art method MvACon \cite{MvACon}. These results further demonstrate the effectiveness of our OcRFDet.

\noindent \textbf{Experiments of other image encoders.}
To further verify the effectiveness of our OcRFDet, we validate our method with small-sized backbone ResNet-18 \cite{he2016resnet} and large-sized backbone SwinTransformer-base \cite{swint} at the same image resolution of 256 $\times$ 704. As shown in Tab. \ref{different image encoder}, when using ResNet-18 as the image backbone, our method achieves 32.9\% mAP and 39.5\% NDS, outperforming our baseline DualBEV by 1.2 pp w.r.t. mAP and 1.8 pp w.r.t. NDS; when using SwinTransformer-base as the image backbone, our method achieves 37.4\% mAP and 41.8\% NDS, outperforming DualBEV by 1.5 pp w.r.t. mAP and 1.7 pp w.r.t. NDS. These results demonstrate the effectiveness of our OcRFDet with different backbones.

\noindent \textbf{Comparison of efficiency.}
We compare the running time, computation cost, and parameter size of the baseline method and our OcRFDet. As shown in Tab. \ref{Comparisons of efficiency}, we find that our OcRFDet brings significant improvements by 1.6 pp w.r.t. mAP and 0.9 pp w.r.t. NDS with only a minimal increase in computational cost. These results demonstrate our method is friendly to applications.

\noindent \textbf{Comparison at different depths.}
Following \cite{cai2023objectfusion}, we categorize annotation and prediction ego distances into three groups: Near (0-20m), Middle (20-30m), and Far ($>$30m). As shown in Tab. \ref{different depths}, compared to the baseline method (DualBEV), our OcRFDet consistently improves performance across all depth ranges. These results indicate our geometric feature enhancement method based on radiance fields is effective across all depth ranges.

    \begin{table}[!h]
        \centering
        \caption{Comparisons at different depths on nuScenes $\tt{validation}$ set. The numbers are \textbf{mAP/NDS}.}
        \setlength{\tabcolsep}{1.7mm}{
        \scalebox{0.9}{
        \begin{tabular}{>{\centering\arraybackslash}m{3.0cm} | >{\centering\arraybackslash}m{1.5cm} >{\centering\arraybackslash}m{1.5cm} >{\centering\arraybackslash}m{1.5cm}} 
        \hline
        Methods & Near & Middle & Far \\
        \hline
        DualBEV \cite{dualbev} & 55.5/53.2 & 26.5/37.4 & 9.8/25.4 \\
        \hline
        \textbf{OcRFDet (Ours)} & \textbf{56.8}/\textbf{53.7} & \textbf{28.3}/\textbf{38.3} & \textbf{10.2}/\textbf{25.9} \\
        \hline
        \end{tabular}
            }
        }
        \label{different depths}
    \end{table}

\noindent \textbf{Comparison at small-sized objects.}
To evaluate the effectiveness of our method for small-sized object detection, we conduct experiments on the nuScenes validation dataset, focusing on normal-sized objects at a far distance ($>$30m) and small-sized objects at a near distance (0-20m). In this context, cars are considered normal-sized objects, while pedestrians, motorcycles, and bicycles are categorized as small-sized objects. As shown in Tab. \ref{Comparisons on small objects}, our method consistently outperforms the baseline DualBEV for the aforementioned small objects. These results indicate that our geometric feature enhancement based on radiance fields significantly improves detection performance for small objects.

    \begin{table}[h]
        \centering
        \begin{minipage}{0.47\textwidth}
            \centering
            \caption{Comparisons at small-sized objects on the nuScenes $\tt{validation}$ set. The numbers are \textbf{AP}.}
            \setlength{\tabcolsep}{1.3mm}{
            \scalebox{0.9}{
            \begin{tabular}{c |c | c c c} 
            \hline
            \multirow{2}{1.5cm}{Methods} & $>$30m & \multicolumn{3}{c}{0-20m} \\
            \cline{2-5}
            & Car & Pedestrian & Motorcycle & Bicycle \\
            \hline
            DualBEV \cite{dualbev} & 23.2 & 55.6 & 53.2 & 48.6 \\ 
            \hline
            \textbf{OcRFDet (Ours)} & \textbf{24.7} & \textbf{55.8} & \textbf{54.7} & \textbf{49.2} \\
            \hline
            \end{tabular} \label{Comparisons on small objects}
                }
            }
        \end{minipage}
    \end{table}

    \begin{table*}[!t]
        \centering
        \caption{Comparisons on the Waymo Open dataset.}
        \label{waymo-mini}
        \setlength{\tabcolsep}{2.0mm}{
        \scalebox{0.9}{
        \begin{tabular}{c | c c | c c | c c } 
        \hline
        \multirow{2}{1.5cm}{Methods} & \multirow{2}{1.5cm}{Backbone} & \multirow{2}{1.5cm}{BEV size} & \multicolumn{2}{c|}{Waymo-full} & \multicolumn{2}{c}{Waymo-mini} \\
        \cline{4-7}
        & & & LET-mAPL $\uparrow$ & LET-mAPH $\uparrow$ & LET-mAPL $\uparrow$ & LET-mAPH $\uparrow$ \\
        \hline
        BEVFormer \cite{bevformer} & \; \multirow{4}{1.7cm}{ResNet-101} & \; \multirow{4}{1.4cm}{200$\times$200} & 35.0 & 47.1 & 34.9 & 46.3 \\
        MvACon \cite{MvACon} & & & - & - & 35.7 & 47.5 \\
        VoctorFormer \cite{voctorformer} & & & 36.8 & 49.1 & - & - \\ 
        \textbf{BEVFormer + Ours} & & & \textbf{37.1} & \textbf{51.3} & \textbf{37.3} & \textbf{48.5} \\
        \hline
        \end{tabular}
            }
        }
    \end{table*}

    \begin{table*}[!t]
        \centering
        \caption{Experiments of OcRFDet with other image backbones on the nuScenes $\tt{validation}$ set.}
        \label{different image encoder}
        \setlength{\tabcolsep}{2.6mm}{
        \scalebox{0.9}{
        \begin{tabular}{c | c | c c |c c c c c} 
        \hline
        Methods & Backbone & mAP $\uparrow$ & NDS $\uparrow$ & mATE $\downarrow$ & mASE $\downarrow$ & mAOE $\downarrow$ & mAVE $\downarrow$ & mAAE $\downarrow$ \\
        \hline
        DualBEV \cite{dualbev} & ResNet-18 & 31.7 & 37.7 & 0.681 & 0.294 & 0.614 & 0.933 & 0.256 \\
        \textbf{OcRFDet (Ours)} & ResNet-18 & \textbf{32.9} & \textbf{39.5} & 0.661 & 0.272 & 0.616 & 0.907 & 0.240 \\
        \hline
        DualBEV \cite{dualbev} & SwinT-Base & 35.9 & 40.1 & 0.677 & 0.269 & 0.534 & 0.993 & 0.312 \\
        \textbf{OcRFDet (Ours)} & SwinT-Base & \textbf{37.4} & \textbf{41.8} & 0.648 & 0.266 & 0.513 & 0.978 & 0.418 \\
        \hline
        \end{tabular}
            }
        }
    \end{table*}
    \begin{table*}[!t]
        \centering
        \caption{Comparisons of running time, computation cost, and parameter size. For a fair comparison, the running speed of the comparison method is evaluated on one RTX 3090 with a batch size of 1.}
        \setlength{\tabcolsep}{3.1mm}{
        \scalebox{0.9}{
        \begin{tabular}{c | c | c c | c c c} 
        \hline
        Methods & Frames & mAP $\uparrow$ & NDS $\uparrow$ & Running Times (FPS) & FLOPs (G) & Parameters (M) \\
        \hline
        DualBEV \cite{dualbev} & 1 & 35.2 & 42.5 & 10.7 & 291.51 & 82.86 \\ 
        \textbf{OcRFDet (Ours)} & 1 & 36.8 & 43.4 & 8.9 & 302.61 & 83.62 \\
        \hline
        DualBEV \cite{dualbev} & 2 & 38.0 & 50.4 & 8.2 & 303.05 & 83.38 \\
        \textbf{OcRFDet (Ours)} & 2 & 40.0 & 50.9 & 6.3 & 314.17 & 84.14 \\
        \hline
        \end{tabular} \label{Comparisons of efficiency}
            }
        }
    \end{table*}

\section{More Ablation Studies} \label{more ablation}

\noindent \textbf{Effectiveness of cross-attention fusion.}
As shown in Fig. \ref{opacity fusion strategy}, among the evaluated strategies, cross-attention fusion achieves the highest detection performance of 35.5\% mAP and 42.2\% NDS, outperforming the other strategies. We think that, compared to linear fusion (weighted mean) or equal-weight fusion (concatenation and convolution), cross-attention fusion further captures more fine-grained associations between the two fields we used, effectively leveraging geometric information. These results demonstrate that cross-attention enables more effective geometry interactions, leading to detection performance improvements.

    \begin{table}[t]
        \centering
            \caption{Ablation studies of the strategy of opacity fusion.}
            \centering
            \setlength{\tabcolsep}{1.3mm}{
            \scalebox{0.9}{
            \begin{tabular}{ >{\centering\arraybackslash}m{5cm} | >{\centering\arraybackslash}m{1.5cm} >{\centering\arraybackslash}m{1.5cm}} 
            \hline
            Strategy & mAP $\uparrow$ & NDS $\uparrow$ \\
            \hline
            Weighted Mean & 35.3 & 41.9 \\
            Concatenation and Convolution & 35.0 & 42.0 \\
            Cross Attention & \textbf{35.5} & \textbf{42.2} \\
            \hline
            \end{tabular}
                }
            }
            \label{opacity fusion strategy}
    \end{table}

\noindent \textbf{Effectiveness of multi-scale height slice attention.}
As shown in Tab. \ref{multi-scale height slice attention}, when we progressively introduce multi-scale integration and BEV mask prediction into HSA, the detection performance improves step by step. These results demonstrate the rationality and effectiveness of our multi-scale height slice attention.

    \begin{figure*}[!htbp]
    \centering
        \begin{minipage}{1\textwidth}
            \centering
            \includegraphics[width=16.0cm]{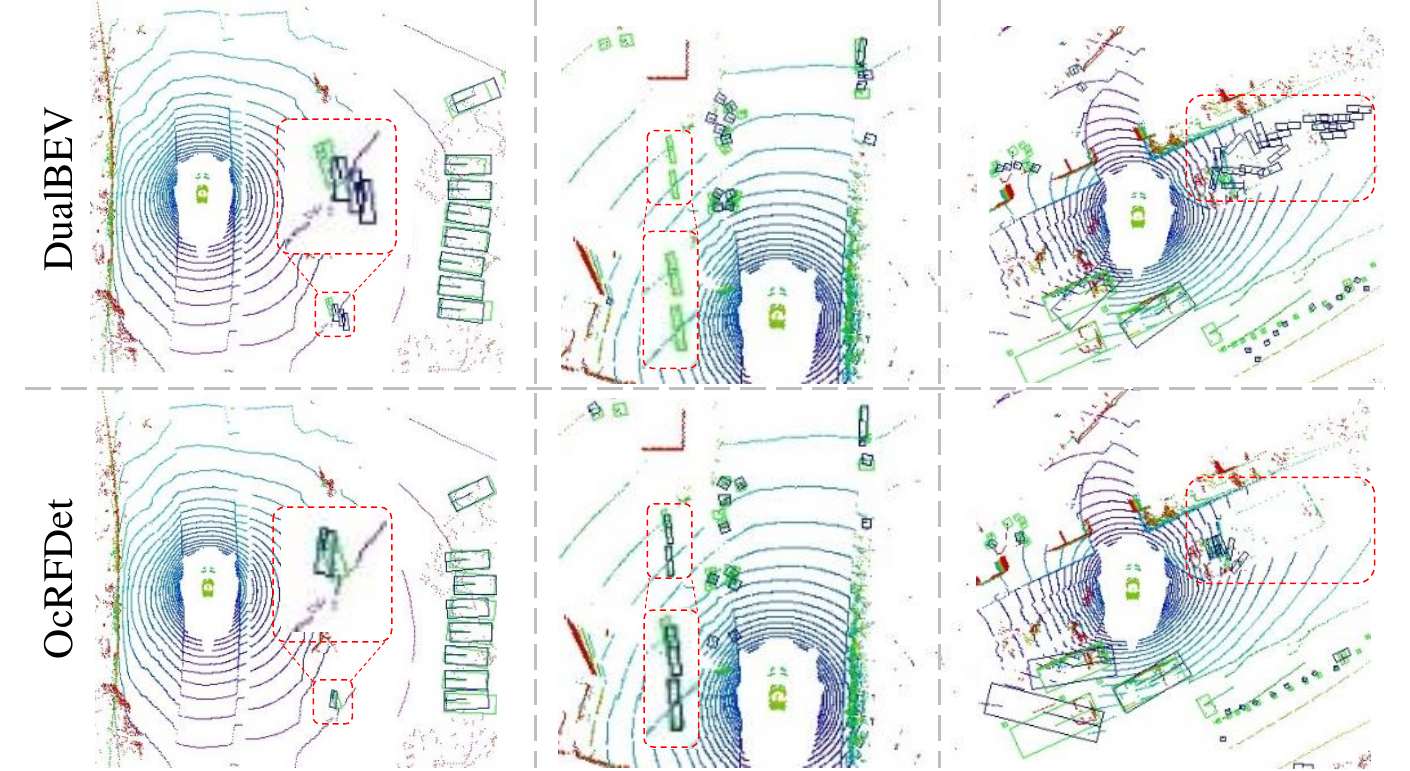}
            \caption{Comparison of detection results in the BEV space on the nuScenes $\tt{validation}$ set. We show the ground truth boxes in \textcolor{green}{green}, and the prediction boxes in \textcolor{blue}{blue}. We use red rectangles to highlight the comparisons of ours and DualBEV.} \label{more visual}
        \end{minipage}
    \end{figure*}

\noindent \textbf{Gains of HOA}
The gains of HOA depend on scene complexity. As shown the Tab. \ref{HOA gains}, scene {\small \texttt{e809$\cdots$eb64}} has only one object category (\textit{Car}), yielding limited improvement (+0.1). In contrast, scene {\small \texttt{3dd2$\cdots$6a0a}} contains seven diverse object types, enabling better use of height-wise distinctions and resulting in a larger gain (\textbf{+1.6}) on \textit{Car}. This shows HOA is more effective in complex scenes.

    \begin{table}[t]
        \centering
            \caption{HOA gains on \textit{Car} under different scene IDs (AP).}
            \centering
            \setlength{\tabcolsep}{1.3mm}{
            \scalebox{0.77}{
            \begin{tabular}{ c | c  c | c} 
            \hline 
            Methods & ID: \texttt{e809$\cdots$eb64} & ID: \texttt{3dd2$\cdots$6a0a} & All scene \\
            \hline
            DualBEV + OcRF & 44.6 & 66.1 & 59.0 \\
            + HOA & \quad \ \ 44.7$^{\uparrow0.1}$  & \quad \ \ 67.7\textbf{$^{\uparrow1.6}$} & \quad \ \ 59.8$^{\uparrow0.8}$  \\
            \hline
            \end{tabular}
                }
            }
            \label{HOA gains}
    \end{table}

\noindent \textbf{Effectiveness of random-view rendering.}
As shown in Tab. \ref{rendering view selection}, we conduct ablations to validate the rationality of randomly selecting a single viewpoint for rendering. Specifically, when rendering all six viewpoints, we find that the detection performance achieves 34.5\% mAP and 41.7\% NDS, with no significant improvement. When rendering four fixed viewpoints (FRONT, FRONT\_LEFT, BACK, and BACK\_RIGHT), the computational cost is reduced, but the detection performance drops by 0.1 pp w.r.t. mAP and 0.3 pp w.r.t. NDS. When rendering four random viewpoints, the detection performance improves by 0.3 pp w.r.t. mAP and 0.3 pp w.r.t. NDS. Furthermore, as fewer viewpoints are randomly selected for rendering, the detection performance further improves, and the computational cost progressively decreases. We analyze these results as follows: Rendering all six viewpoints focuses the network’s optimization on the radiance fields, limiting the optimization of the detection network while incurring high computational costs. Rendering randomly selected viewpoints ensures that all six viewpoints are rendered throughout the training process, making it more beneficial for detection compared to rendering the same number of fixed viewpoints. Finally, to balance detection performance and computational efficiency, we identify that randomly rendering a single viewpoint is optimal.

    \begin{table}[t]
        \centering
            \caption{Ablation studies of multi-scale height slice attention.}
            \centering
            \setlength{\tabcolsep}{1.7mm}{
            \scalebox{0.9}{
            \begin{tabular}{ >{\centering\arraybackslash}m{1.4cm} >{\centering\arraybackslash}m{1.7cm} >{\centering\arraybackslash}m{1.7cm} | >{\centering\arraybackslash}m{1.2cm} >{\centering\arraybackslash}m{1.2cm}} 
            \hline
            HSA & Multi Scale & BEV Mask & mAP $\uparrow$ & NDS $\uparrow$ \\
            \hline
            \ding{51} & & & 35.2 & 42.0 \\
            \ding{51} & \ding{51} & & 35.3 & 42.1 \\
            \ding{51} & \ding{51} & \ding{51} & \textbf{35.5} & \textbf{42.2} \\
            \hline
            \end{tabular}
                }
            }
            \label{multi-scale height slice attention}
    \end{table}

    \begin{table}[!h]
        \centering
            \caption{Ablation studies of rendering view selection.}
            \centering
            \setlength{\tabcolsep}{1.3mm}{
            \scalebox{0.9}{
            \begin{tabular}{ c | c c | c c} 
            \hline
            \multirow{2}{0.8cm}{View} & \multirow{2}{0.8cm}{mAP}& \multirow{2}{0.8cm}{NDS} & GPU Memory & Training Time of \\
            & & & (G) & Each Iteration (s) \\
            \hline
            6 & 34.5 & 41.7 & 23.9 & 0.744 \\
            4 (Fixed) & 34.4 & 41.4 & 19.2 & 0.625 \\
            4 (Random) & 34.7 & 41.7 & 19.2 & 0.635 \\
            2 (Random) & 34.7 & \textbf{42.1} & 16.9 & 0.527 \\
            1 (Random) & \textbf{35.0} & 41.9 & 12.5 & 0.459 \\
            \hline
            \end{tabular}
                }
            }
            \label{rendering view selection}
    \end{table}

\section{More Qualitative Results} \label{more qualitative}
We further compare the qualitative results of DualBEV and ours in the BEV space on the nuScenes $\tt{validation}$ set. As shown in Fig. \ref{more visual}, in the left column, our method shows a more accurate location of the predicted boxes for the distant objects. In the middle column, our method successfully obtains the detection boxes for occluded objects. In the right column, our method produces fewer false positive boxes. These results further demonstrate the superiority of our OcRFDet.

\section{Limitation and Future Work} \label{limitation}
While our object-centric radiance fields effectively enhance foreground objects and suppress background noise, the learning process is heavily dependent on the quality of annotations. This leads to two interrelated constraints: (1) Critical foreground instances that lack 3D box labels (e.g., traffic lights in the nuScenes dataset) may be inadvertently suppressed during feature enhancement, which limits the applicability of our method in open environments. (2) Inaccurate 3D box labels in the training data could lead our method to unintentionally amplify background features that should have been suppressed. Future work will explore leveraging prompts as priors in open-set datasets to preserve important yet unlabeled foreground object features, while addressing label noise through temporal consistency constraints in dynamic scenes.

{
    \small
    \bibliographystyle{ieeenat_fullname}
    \bibliography{main}

\begin{thebibliography}{61}
\providecommand{\natexlab}[1]{#1}
\providecommand{\url}[1]{\texttt{#1}}
\expandafter\ifx\csname urlstyle\endcsname\relax
  \providecommand{\doi}[1]{doi: #1}\else
  \providecommand{\doi}{doi: \begingroup \urlstyle{rm}\Url}\fi

\bibitem[Caesar et~al.(2020)Caesar, Bankiti, Lang, Vora, Liong, Xu, Krishnan, Pan, Baldan, and Beijbom]{caesar2020nuscenes}
Holger Caesar, Varun Bankiti, Alex~H Lang, Sourabh Vora, Venice~Erin Liong, Qiang Xu, Anush Krishnan, Yu Pan, Giancarlo Baldan, and Oscar Beijbom.
\newblock nuscenes: A multimodal dataset for autonomous driving.
\newblock In \emph{CVPR}, 2020.

\bibitem[Cai et~al.(2023)Cai, Pan, Yao, Ngo, and Mei]{cai2023objectfusion}
Qi Cai, Yingwei Pan, Ting Yao, Chong-Wah Ngo, and Tao Mei.
\newblock Objectfusion: Multi-modal 3d object detection with object-centric fusion.
\newblock In \emph{ICCV}, 2023.

\bibitem[Chabot et~al.(2024)Chabot, Granger, and Lapouge]{gaussianbev}
Florian Chabot, Nicolas Granger, and Guillaume Lapouge.
\newblock Gaussianbev: 3d gaussian representation meets perception models for bev segmentation.
\newblock \emph{arXiv preprint arXiv:2407.14108}, 2024.

\bibitem[Chang et~al.(2024)Chang, Zhang, Zhang, Zhao, He, and Liu]{recurrentbev}
Ming Chang, Xishan Zhang, Rui Zhang, Zhipeng Zhao, Guanhua He, and Shaoli Liu.
\newblock Recurrentbev: a long-term temporal fusion framework for multi-view 3d detection.
\newblock In \emph{ECCV}, 2024.

\bibitem[Chen et~al.(2024)Chen, Xu, Ye, Qian, Zou, Yeung, and Chen]{voctorformer}
Zhili Chen, Shuangjie Xu, Maosheng Ye, Zian Qian, Xiaoyi Zou, Dit-Yan Yeung, and Qifeng Chen.
\newblock Learning high-resolution vector representation from multi-camera images for 3d object detection.
\newblock In \emph{ECCV}, 2024.

\bibitem[Chi et~al.(2023)Chi, Liu, Lu, Zhang, Wang, Guo, and Zhang]{bevsan}
Xiaowei Chi, Jiaming Liu, Ming Lu, Rongyu Zhang, Zhaoqing Wang, Yandong Guo, and Shanghang Zhang.
\newblock Bev-san: Accurate bev 3d object detection via slice attention networks.
\newblock In \emph{CVPR}, 2023.

\bibitem[Chung et~al.(2024)Chung, Oh, and Lee]{depth3dgs}
Jaeyoung Chung, Jeongtaek Oh, and Kyoung~Mu Lee.
\newblock Depth-regularized optimization for 3d gaussian splatting in few-shot images.
\newblock In \emph{CVPR}, 2024.

\bibitem[Contributors(2020)]{mmdet3d2020}
MMDetection3D Contributors.
\newblock {MMDetection3D: OpenMMLab} next-generation platform for general {3D} object detection.
\newblock \url{https://github.com/open-mmlab/mmdetection3d}, 2020.

\bibitem[Han et~al.(2024)Han, Yang, Sun, Ge, Dong, Zhou, Mao, Peng, and Zhang]{videobev}
Chunrui Han, Jinrong Yang, Jianjian Sun, Zheng Ge, Runpei Dong, Hongyu Zhou, Weixin Mao, Yuang Peng, and Xiangyu Zhang.
\newblock Exploring recurrent long-term temporal fusion for multi-view 3d perception.
\newblock \emph{IEEE RA-L}, 2024.

\bibitem[He et~al.(2016)He, Zhang, Ren, and Sun]{he2016resnet}
Kaiming He, Xiangyu Zhang, Shaoqing Ren, and Jian Sun.
\newblock Deep residual learning for image recognition.
\newblock In \emph{CVPR}, 2016.

\bibitem[Hou et~al.(2024)Hou, Wang, Ye, Liu, Gong, Tan, Ding, Wang, and Bai]{open}
Jinghua Hou, Tong Wang, Xiaoqing Ye, Zhe Liu, Shi Gong, Xiao Tan, Errui Ding, Jingdong Wang, and Xiang Bai.
\newblock Open: Object-wise position embedding for multi-view 3d object detection.
\newblock In \emph{ECCV}, 2024.

\bibitem[Huang and Huang(2022)]{bevdet4d}
Junjie Huang and Guan Huang.
\newblock Bevdet4d: Exploit temporal cues in multi-camera 3d object detection.
\newblock \emph{arXiv preprint arXiv:2203.17054}, 2022.

\bibitem[Huang et~al.(2021)Huang, Huang, Zhu, Ye, and Du]{bevdet}
Junjie Huang, Guan Huang, Zheng Zhu, Yun Ye, and Dalong Du.
\newblock Bevdet: High-performance multi-camera 3d object detection in bird-eye-view.
\newblock \emph{arXiv preprint arXiv:2112.11790}, 2021.

\bibitem[Huang et~al.(2024)Huang, Zheng, Zhang, Zhou, and Lu]{gaussianformer}
Yuanhui Huang, Wenzhao Zheng, Yunpeng Zhang, Jie Zhou, and Jiwen Lu.
\newblock Gaussianformer: Scene as gaussians for vision-based 3d semantic occupancy prediction.
\newblock \emph{arXiv preprint arXiv:2405.17429}, 2024.

\bibitem[Ji et~al.(2024)Ji, Liang, and Cheng]{QAF2D}
Haoxuanye Ji, Pengpeng Liang, and Erkang Cheng.
\newblock Enhancing 3d object detection with 2d detection-guided query anchors.
\newblock In \emph{CVPR}, 2024.

\bibitem[Kerbl et~al.(2023)Kerbl, Kopanas, Leimk{\"u}hler, and Drettakis]{3DGS}
Bernhard Kerbl, Georgios Kopanas, Thomas Leimk{\"u}hler, and George Drettakis.
\newblock 3d gaussian splatting for real-time radiance field rendering.
\newblock \emph{ACM Trans. Graph.}, 2023.

\bibitem[Kingma and Ba(2014)]{adamw}
Diederik~P Kingma and Jimmy Ba.
\newblock Adam: A method for stochastic optimization.
\newblock \emph{arXiv preprint arXiv:1412.6980}, 2014.

\bibitem[Kumar et~al.(2024)Kumar, Guo, Huang, Ren, and Liu]{seabird}
Abhinav Kumar, Yuliang Guo, Xinyu Huang, Liu Ren, and Xiaoming Liu.
\newblock Seabird: Segmentation in bird's view with dice loss improves monocular 3d detection of large objects.
\newblock In \emph{CVPR}, 2024.

\bibitem[Li et~al.(2023{\natexlab{a}})Li, Sima, Dai, Wang, Lu, Wang, Zeng, Li, Yang, Deng, et~al.]{li2023delving}
Hongyang Li, Chonghao Sima, Jifeng Dai, Wenhai Wang, Lewei Lu, Huijie Wang, Jia Zeng, Zhiqi Li, Jiazhi Yang, Hanming Deng, et~al.
\newblock Delving into the devils of bird's-eye-view perception: A review, evaluation and recipe.
\newblock \emph{IEEE TPAMI}, 2023{\natexlab{a}}.

\bibitem[Li et~al.(2024{\natexlab{a}})Li, Shen, Huang, and Cui]{dualbev}
Peidong Li, Wancheng Shen, Qihao Huang, and Dixiao Cui.
\newblock Dualbev: Cnn is all you need in view transformation.
\newblock In \emph{ECCV}, 2024{\natexlab{a}}.

\bibitem[Li et~al.(2023{\natexlab{b}})Li, Bao, Ge, Yang, Sun, and Li]{bevstereo}
Yinhao Li, Han Bao, Zheng Ge, Jinrong Yang, Jianjian Sun, and Zeming Li.
\newblock Bevstereo: Enhancing depth estimation in multi-view 3d object detection with temporal stereo.
\newblock In \emph{AAAI}, 2023{\natexlab{b}}.

\bibitem[Li et~al.(2023{\natexlab{c}})Li, Ge, Yu, Yang, Wang, Shi, Sun, and Li]{bevdepth}
Yinhao Li, Zheng Ge, Guanyi Yu, Jinrong Yang, Zengran Wang, Yukang Shi, Jianjian Sun, and Zeming Li.
\newblock Bevdepth: Acquisition of reliable depth for multi-view 3d object detection.
\newblock In \emph{AAAI}, 2023{\natexlab{c}}.

\bibitem[Li et~al.(2022)Li, Wang, Li, Xie, Sima, Lu, Qiao, and Dai]{bevformer}
Zhiqi Li, Wenhai Wang, Hongyang Li, Enze Xie, Chonghao Sima, Tong Lu, Yu Qiao, and Jifeng Dai.
\newblock Bevformer: Learning bird’s-eye-view representation from multi-camera images via spatiotemporal transformers.
\newblock In \emph{ECCV}, 2022.

\bibitem[Li et~al.(2023{\natexlab{d}})Li, Yu, Wang, Anandkumar, Lu, and Alvarez]{fbbev}
Zhiqi Li, Zhiding Yu, Wenhai Wang, Anima Anandkumar, Tong Lu, and Jose~M Alvarez.
\newblock Fb-bev: Bev representation from forward-backward view transformations.
\newblock In \emph{ICCV}, 2023{\natexlab{d}}.

\bibitem[Li et~al.(2024{\natexlab{b}})Li, Lan, Alvarez, and Wu]{bevnext}
Zhenxin Li, Shiyi Lan, Jose~M Alvarez, and Zuxuan Wu.
\newblock Bevnext: Reviving dense bev frameworks for 3d object detection.
\newblock In \emph{CVPR}, 2024{\natexlab{b}}.

\bibitem[Lin et~al.(2017)Lin, Doll{\'a}r, Girshick, He, Hariharan, and Belongie]{lin2017fpn}
Tsung-Yi Lin, Piotr Doll{\'a}r, Ross Girshick, Kaiming He, Bharath Hariharan, and Serge Belongie.
\newblock Feature pyramid networks for object detection.
\newblock In \emph{CVPR}, 2017.

\bibitem[Lin et~al.(2022)Lin, Lin, Pei, Huang, and Su]{sparse4d}
Xuewu Lin, Tianwei Lin, Zixiang Pei, Lichao Huang, and Zhizhong Su.
\newblock Sparse4d: Multi-view 3d object detection with sparse spatial-temporal fusion.
\newblock \emph{arXiv preprint arXiv:2211.10581}, 2022.

\bibitem[Liu et~al.(2024{\natexlab{a}})Liu, Huang, Zhang, Yao, Zhang, Wan, Ye, and Zhou]{raydn}
Feng Liu, Tengteng Huang, Qianjing Zhang, Haotian Yao, Chi Zhang, Fang Wan, Qixiang Ye, and Yanzhao Zhou.
\newblock Ray denoising: Depth-aware hard negative sampling for multi-view 3d object detection.
\newblock In \emph{ECCV}, 2024{\natexlab{a}}.

\bibitem[Liu et~al.(2023)Liu, Teng, Lu, Wang, and Wang]{sparsebev}
Haisong Liu, Yao Teng, Tao Lu, Haiguang Wang, and Limin Wang.
\newblock Sparsebev: High-performance sparse 3d object detection from multi-camera videos.
\newblock In \emph{ICCV}, 2023.

\bibitem[Liu et~al.(2024{\natexlab{b}})Liu, Wang, Hu, Shen, Ye, Zang, Cao, Li, and Liu]{generalizable3dgs}
Tianqi Liu, Guangcong Wang, Shoukang Hu, Liao Shen, Xinyi Ye, Yuhang Zang, Zhiguo Cao, Wei Li, and Ziwei Liu.
\newblock Fast generalizable gaussian splatting reconstruction from multi-view stereo.
\newblock In \emph{ECCV}, 2024{\natexlab{b}}.

\bibitem[Liu et~al.(2024{\natexlab{c}})Liu, Zheng, Qian, Xue, Chen, Zhang, Li, and Wu]{MvACon}
Xianpeng Liu, Ce Zheng, Ming Qian, Nan Xue, Chen Chen, Zhebin Zhang, Chen Li, and Tianfu Wu.
\newblock Multi-view attentive contextualization for multi-view 3d object detection.
\newblock In \emph{CVPR}, 2024{\natexlab{c}}.

\bibitem[Liu et~al.(2021)Liu, Lin, Cao, Hu, Wei, Zhang, Lin, and Guo]{swint}
Ze Liu, Yutong Lin, Yue Cao, Han Hu, Yixuan Wei, Zheng Zhang, Stephen Lin, and Baining Guo.
\newblock Swin transformer: Hierarchical vision transformer using shifted windows.
\newblock In \emph{ICCV}, 2021.

\bibitem[Liu et~al.(2022)Liu, Mao, Wu, Feichtenhofer, Darrell, and Xie]{convnext}
Zhuang Liu, Hanzi Mao, Chao-Yuan Wu, Christoph Feichtenhofer, Trevor Darrell, and Saining Xie.
\newblock A convnet for the 2020s.
\newblock In \emph{CVPR}, 2022.

\bibitem[Martin-Brualla et~al.(2021)Martin-Brualla, Radwan, Sajjadi, Barron, Dosovitskiy, and Duckworth]{nerfwild}
Ricardo Martin-Brualla, Noha Radwan, Mehdi~SM Sajjadi, Jonathan~T Barron, Alexey Dosovitskiy, and Daniel Duckworth.
\newblock Nerf in the wild: Neural radiance fields for unconstrained photo collections.
\newblock In \emph{CVPR}, 2021.

\bibitem[Mildenhall et~al.(2021)Mildenhall, Srinivasan, Tancik, Barron, Ramamoorthi, and Ng]{nerf}
Ben Mildenhall, Pratul~P Srinivasan, Matthew Tancik, Jonathan~T Barron, Ravi Ramamoorthi, and Ren Ng.
\newblock Nerf: Representing scenes as neural radiance fields for view synthesis.
\newblock \emph{Communications of the ACM}, 2021.

\bibitem[Milletari et~al.(2016)Milletari, Navab, and Ahmadi]{diceloss}
Fausto Milletari, Nassir Navab, and Seyed-Ahmad Ahmadi.
\newblock V-net: Fully convolutional neural networks for volumetric medical image segmentation.
\newblock In \emph{3DV}, 2016.

\bibitem[Pan et~al.(2023)Pan, Liu, Liu, Huang, Wang, Zhang, Xu, Lai, and Yang]{uniocc}
Mingjie Pan, Li Liu, Jiaming Liu, Peixiang Huang, Longlong Wang, Shanghang Zhang, Shaoqing Xu, Zhiyi Lai, and Kuiyuan Yang.
\newblock Uniocc: Unifying vision-centric 3d occupancy prediction with geometric and semantic rendering.
\newblock \emph{arXiv preprint arXiv:2306.09117}, 2023.

\bibitem[Pan et~al.(2024)Pan, Liu, Zhang, Huang, Li, Xie, Wang, Liu, and Zhang]{renderocc}
Mingjie Pan, Jiaming Liu, Renrui Zhang, Peixiang Huang, Xiaoqi Li, Hongwei Xie, Bing Wang, Li Liu, and Shanghang Zhang.
\newblock Renderocc: Vision-centric 3d occupancy prediction with 2d rendering supervision.
\newblock In \emph{ICRA}, 2024.

\bibitem[Park et~al.(2021)Park, Ambrus, Guizilini, Li, and Gaidon]{dd3d}
Dennis Park, Rares Ambrus, Vitor Guizilini, Jie Li, and Adrien Gaidon.
\newblock Is pseudo-lidar needed for monocular 3d object detection?
\newblock In \emph{ICCV}, 2021.

\bibitem[Park et~al.(2022)Park, Xu, Yang, Keutzer, Kitani, Tomizuka, and Zhan]{solofusion}
Jinhyung Park, Chenfeng Xu, Shijia Yang, Kurt Keutzer, Kris~M Kitani, Masayoshi Tomizuka, and Wei Zhan.
\newblock Time will tell: New outlooks and a baseline for temporal multi-view 3d object detection.
\newblock In \emph{ICLR}, 2022.

\bibitem[Paszke et~al.(2019)Paszke, Gross, Massa, Lerer, Bradbury, Chanan, Killeen, Lin, Gimelshein, Antiga, et~al.]{paszke2019pytorch}
Adam Paszke, Sam Gross, Francisco Massa, Adam Lerer, James Bradbury, Gregory Chanan, Trevor Killeen, Zeming Lin, Natalia Gimelshein, Luca Antiga, et~al.
\newblock Pytorch: An imperative style, high-performance deep learning library.
\newblock In \emph{NeurIPS}, 2019.

\bibitem[Philion and Fidler(2020)]{lss}
Jonah Philion and Sanja Fidler.
\newblock Lift, splat, shoot: Encoding images from arbitrary camera rigs by implicitly unprojecting to 3d.
\newblock In \emph{ECCV}, 2020.

\bibitem[Sun et~al.(2020)Sun, Kretzschmar, Dotiwalla, Chouard, Patnaik, Tsui, Guo, Zhou, Chai, Caine, et~al.]{Waymo}
Pei Sun, Henrik Kretzschmar, Xerxes Dotiwalla, Aurelien Chouard, Vijaysai Patnaik, Paul Tsui, James Guo, Yin Zhou, Yuning Chai, Benjamin Caine, et~al.
\newblock Scalability in perception for autonomous driving: Waymo open dataset.
\newblock In \emph{CVPR}, 2020.

\bibitem[Tancik et~al.(2022)Tancik, Casser, Yan, Pradhan, Mildenhall, Srinivasan, Barron, and Kretzschmar]{blocknerf}
Matthew Tancik, Vincent Casser, Xinchen Yan, Sabeek Pradhan, Ben Mildenhall, Pratul~P Srinivasan, Jonathan~T Barron, and Henrik Kretzschmar.
\newblock Block-nerf: Scalable large scene neural view synthesis.
\newblock In \emph{CVPR}, 2022.

\bibitem[Wang et~al.(2023)Wang, Liu, Wang, Li, and Zhang]{streampetr}
Shihao Wang, Yingfei Liu, Tiancai Wang, Ying Li, and Xiangyu Zhang.
\newblock Exploring object-centric temporal modeling for efficient multi-view 3d object detection.
\newblock In \emph{ICCV}, 2023.

\bibitem[Wang et~al.(2021)Wang, Zhu, Pang, and Lin]{v299}
Tai Wang, Xinge Zhu, Jiangmiao Pang, and Dahua Lin.
\newblock Fcos3d: Fully convolutional one-stage monocular 3d object detection.
\newblock In \emph{ICCV}, 2021.

\bibitem[Wang et~al.(2022)Wang, Guizilini, Zhang, Wang, Zhao, and Solomon]{detr3d}
Yue Wang, Vitor~Campagnolo Guizilini, Tianyuan Zhang, Yilun Wang, Hang Zhao, and Justin Solomon.
\newblock Detr3d: 3d object detection from multi-view images via 3d-to-2d queries.
\newblock In \emph{RL}, 2022.

\bibitem[Wang et~al.(2004)Wang, Bovik, Sheikh, and Simoncelli]{ssim}
Zhou Wang, Alan~C Bovik, Hamid~R Sheikh, and Eero~P Simoncelli.
\newblock Image quality assessment: from error visibility to structural similarity.
\newblock \emph{IEEE TIP}, 2004.

\bibitem[Wu et~al.(2023)Wu, Sun, Shen, and Zhang]{mapnerf}
Chenming Wu, Jiadai Sun, Zhelun Shen, and Liangjun Zhang.
\newblock Mapnerf: Incorporating map priors into neural radiance fields for driving view simulation.
\newblock In \emph{IROS}, 2023.

\bibitem[Wu et~al.(2024{\natexlab{a}})Wu, Yi, Fang, Xie, Zhang, Wei, Liu, Tian, and Wang]{4DGS}
Guanjun Wu, Taoran Yi, Jiemin Fang, Lingxi Xie, Xiaopeng Zhang, Wei Wei, Wenyu Liu, Qi Tian, and Xinggang Wang.
\newblock 4d gaussian splatting for real-time dynamic scene rendering.
\newblock In \emph{CVPR}, 2024{\natexlab{a}}.

\bibitem[Wu et~al.(2024{\natexlab{b}})Wu, Li, Qin, Zhao, and Li]{heightformer}
Yiming Wu, Ruixiang Li, Zequn Qin, Xinhai Zhao, and Xi Li.
\newblock Heightformer: Explicit height modeling without extra data for camera-only 3d object detection in bird’s eye view.
\newblock \emph{IEEE TIP}, 2024{\natexlab{b}}.

\bibitem[Xu et~al.(2023)Xu, Wu, Hou, Tsai, Li, Wang, Zhan, He, Vajda, Keutzer, et~al.]{nerfdet}
Chenfeng Xu, Bichen Wu, Ji Hou, Sam Tsai, Ruilong Li, Jialiang Wang, Wei Zhan, Zijian He, Peter Vajda, Kurt Keutzer, et~al.
\newblock Nerf-det: Learning geometry-aware volumetric representation for multi-view 3d object detection.
\newblock In \emph{ICCV}, 2023.

\bibitem[Yan et~al.(2024)Yan, Lin, Zhou, Wang, Sun, Zhan, Lang, Zhou, and Peng]{streetgaussian}
Yunzhi Yan, Haotong Lin, Chenxu Zhou, Weijie Wang, Haiyang Sun, Kun Zhan, Xianpeng Lang, Xiaowei Zhou, and Sida Peng.
\newblock Street gaussians for modeling dynamic urban scenes.
\newblock \emph{arXiv preprint arXiv:2401.01339}, 2024.

\bibitem[Yang et~al.(2023{\natexlab{a}})Yang, Chen, Tian, Tao, Zhu, Zhang, Huang, Li, Qiao, Lu, et~al.]{bevformerv2}
Chenyu Yang, Yuntao Chen, Hao Tian, Chenxin Tao, Xizhou Zhu, Zhaoxiang Zhang, Gao Huang, Hongyang Li, Yu Qiao, Lewei Lu, et~al.
\newblock Bevformer v2: Adapting modern image backbones to bird's-eye-view recognition via perspective supervision.
\newblock In \emph{CVPR}, 2023{\natexlab{a}}.

\bibitem[Yang et~al.(2023{\natexlab{b}})Yang, Ivanovic, Litany, Weng, Kim, Li, Che, Xu, Fidler, Pavone, et~al.]{emernerf}
Jiawei Yang, Boris Ivanovic, Or Litany, Xinshuo Weng, Seung~Wook Kim, Boyi Li, Tong Che, Danfei Xu, Sanja Fidler, Marco Pavone, et~al.
\newblock Emernerf: Emergent spatial-temporal scene decomposition via self-supervision.
\newblock \emph{arXiv preprint arXiv:2311.02077}, 2023{\natexlab{b}}.

\bibitem[Yin et~al.(2021)Yin, Zhou, and Krahenbuhl]{centerpoint}
Tianwei Yin, Xingyi Zhou, and Philipp Krahenbuhl.
\newblock Center-based 3d object detection and tracking.
\newblock In \emph{CVPR}, 2021.

\bibitem[Zhang et~al.(2023)Zhang, Zhang, Liu, and Wang]{sabev}
Jinqing Zhang, Yanan Zhang, Qingjie Liu, and Yunhong Wang.
\newblock Sa-bev: Generating semantic-aware bird's-eye-view feature for multi-view 3d object detection.
\newblock In \emph{ICCV}, 2023.

\bibitem[Zhang et~al.(2024)Zhang, Zhang, Kuang, and Zhang]{nerflidar}
Junge Zhang, Feihu Zhang, Shaochen Kuang, and Li Zhang.
\newblock Nerf-lidar: Generating realistic lidar point clouds with neural radiance fields.
\newblock In \emph{AAAI}, 2024.

\bibitem[Zhao et~al.(2024)Zhao, Chen, Sun, Yang, Wang, Zhang, Li, Kou, Wei, and Zhang]{hybridocc}
Xiao Zhao, Bo Chen, Mingyang Sun, Dingkang Yang, Youxing Wang, Xukun Zhang, Mingcheng Li, Dongliang Kou, Xiaoyi Wei, and Lihua Zhang.
\newblock Hybridocc: Nerf enhanced transformer-based multi-camera 3d occupancy prediction.
\newblock \emph{IEEE RA-L}, 2024.

\bibitem[Zhou et~al.(2024{\natexlab{a}})Zhou, Shao, Xu, Bai, Qiu, Liu, Wang, Geiger, and Liao]{hugs}
Hongyu Zhou, Jiahao Shao, Lu Xu, Dongfeng Bai, Weichao Qiu, Bingbing Liu, Yue Wang, Andreas Geiger, and Yiyi Liao.
\newblock Hugs: Holistic urban 3d scene understanding via gaussian splatting.
\newblock In \emph{CVPR}, 2024{\natexlab{a}}.

\bibitem[Zhou et~al.(2024{\natexlab{b}})Zhou, Lin, Shan, Wang, Sun, and Yang]{drivinggaussian}
Xiaoyu Zhou, Zhiwei Lin, Xiaojun Shan, Yongtao Wang, Deqing Sun, and Ming-Hsuan Yang.
\newblock Drivinggaussian: Composite gaussian splatting for surrounding dynamic autonomous driving scenes.
\newblock In \emph{CVPR}, 2024{\natexlab{b}}.

\end{thebibliography}
}

\end{document}